%% file: acl2023.tex
\pdfoutput=1

\documentclass[11pt]{article}

\usepackage[]{ACL2023}

\usepackage{times}
\usepackage{latexsym}

\usepackage[T1]{fontenc}

\usepackage[utf8]{inputenc}

\usepackage{microtype}

\usepackage{inconsolata}

\usepackage{xcolor}
\usepackage{todonotes}
\usepackage{xspace}

\usepackage{amsmath}
\usepackage{mathtools}
\usepackage{cleveref}
\usepackage{fdsymbol}
\usepackage[normalem]{ulem}
\usepackage{booktabs}
\usepackage{adjustbox}
\usepackage{colortbl}
\usepackage[inline]{enumitem}
\usepackage{wrapfig}
\usepackage{multirow}

\usepackage{xcolor}
\definecolor{olmocolor}{HTML}{60D2FB}

\newcommand{\dolma}{Dolma\xspace}
\newcommand{\olmo}{OLMo\xspace}
\newcommand{\model}{OLMo-7B\xspace}
\newcommand{\modelsmall}{OLMo-1B\xspace}
\newcommand{\tulu}{\textsc{T\"ulu}\xspace}
\newcommand{\instructgpt}[1]{\text{Davinci}-{\text{#1}}}

\newcommand{\olmoLogoWithText}{\raisebox{-.3em}{\rlap{\raisebox{.3em}{\hspace{1.4em}\scriptsize OLMo}}\includegraphics[height=1.5em]{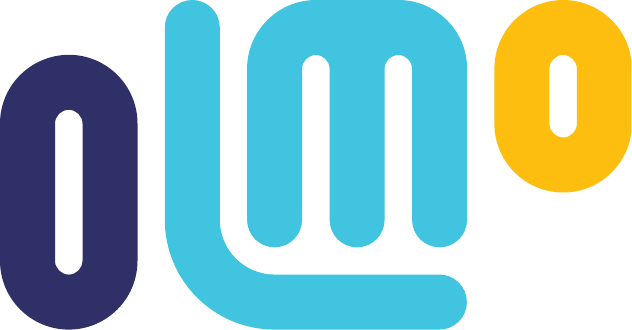}}\xspace}

\definecolor{olmoDarkBlue}{HTML}{012e59}
\definecolor{olmoBlue}{HTML}{265ed4}
\definecolor{olmoLightBlue}{HTML}{012e59}
\definecolor{olmoTeal}{HTML}{00d5ff}
\definecolor{olmoYellow}{HTML}{ffbb00}
\definecolor{olmoOrange}{HTML}{ff9100}

\definecolor{tableGray}{HTML}{C0C0C0}
\newcolumntype{g}{>{\columncolor{tableGray}}c}

\newcommand{\balpha}{{\color{olmoBlue}\boldsymbol{\alpha}}}
\newcommand{\bbeta}{{\color{olmoBlue}\boldsymbol{\beta}}}
\newcommand{\bgamma}{{\color{olmoBlue}\boldsymbol{\gamma}}}
\newcommand{\bdelta}{{\color{olmoBlue}\boldsymbol{\delta}}}
\newcommand{\bmu}{{\color{olmoBlue}\boldsymbol{\mu}}}

\title{\olmoLogoWithText~: Accelerating the Science of Language Models}

\author{
  {\bf
    Dirk Groeneveld
    \hspace{-0.5em}
    $^{\balpha}$
    \enskip
    Iz Beltagy
    \hspace{-0.5em}
    $^{\balpha}$
   \vspace{8pt}
  } \\
  {\bf
    Pete Walsh
    \hspace{-0.5em}
    $^{\balpha}$
    \enskip
    Akshita Bhagia
    \hspace{-0.5em}
    $^{\balpha}$
    \enskip
    Rodney Kinney
    \hspace{-0.5em}
    $^{\balpha}$
    \enskip
    Oyvind Tafjord
    \hspace{-0.5em}
    $^{\balpha}$
    \vspace{8pt}
  } \\
  {\bf
    Ananya Harsh Jha
    \hspace{-0.5em}
    $^{\balpha}$
    \enskip
    Hamish Ivison
    \hspace{-0.5em}
    $^{\balpha}$$^{\bbeta}$
    \enskip
    Ian Magnusson
    \hspace{-0.5em}
    $^{\balpha}$
    \enskip
    Yizhong Wang
    \hspace{-0.5em}
    $^{\balpha}$$^{\bbeta}$
    \vspace{8pt}
  } \\
  {\bf
    Shane Arora
    \hspace{-0.5em}
    $^{\balpha}$
    \enskip
    David Atkinson
    \hspace{-0.5em}
    $^{\balpha}$
    \enskip
    Russell Authur
    \hspace{-0.5em}
    $^{\balpha}$
    \enskip
    Khyathi Raghavi Chandu
    \hspace{-0.5em}
    $^{\balpha}$
    \vspace{1.5pt}
  } \\
  {\bf
    Arman Cohan
    \hspace{-0.5em}
    $^{\bgamma}$$^{\balpha}$
    \enskip
    Jennifer Dumas
    \hspace{-0.5em}
    $^{\balpha}$
    \enskip
    Yanai Elazar
    \hspace{-0.5em}
    $^{\balpha}$$^{\bbeta}$
    \enskip
    Yuling Gu
    \hspace{-0.5em}
    $^{\balpha}$
    \enskip
    \vspace{1.5pt}
  } \\
  {\bf
    Jack Hessel
    \hspace{-0.5em}
    $^{\balpha}$
    \enskip
    Tushar Khot
    \hspace{-0.5em}
    $^{\balpha}$
    \enskip
    William Merrill
    \hspace{-0.5em}
    $^{\bdelta}$
    \enskip
    Jacob Morrison
    \hspace{-0.5em}
    $^{\balpha}$
    \vspace{1.5pt}
  } \\
  {\bf
    \vspace{1.5pt}
    Niklas Muennighoff
    \hspace{-0.5em}
    $ $
    \enskip
    Aakanksha Naik
    \hspace{-0.5em}
    $^{\balpha}$
    \enskip
    Crystal Nam
    \hspace{-0.5em}
    $^{\balpha}$
    \enskip
    Matthew E. Peters
    \hspace{-0.5em}
    $^{\balpha}$
    \vspace{1.5pt}
  } \\
  {\bf
    Valentina Pyatkin
    \hspace{-0.5em}
    $^{\balpha}$$^{\bbeta}$
    \enskip
    Abhilasha Ravichander
    \hspace{-0.5em}
    $^{\balpha}$
    \enskip
    Dustin Schwenk
    \hspace{-0.5em}
    $^{\balpha}$
    \enskip
    Saurabh Shah
    \hspace{-0.5em}
    $^{\balpha}$
    \enskip
    \vspace{1.5pt}
  } \\
  {\bf
    Will Smith
    \hspace{-0.5em}
    $^{\balpha}$
    \enskip
    Emma Strubell
    \hspace{-0.5em}
    $^{\balpha}$$^{\bmu}$
    \enskip
    Nishant Subramani
    \hspace{-0.5em}
    $^{\balpha}$
    \enskip
    Mitchell Wortsman
    \hspace{-0.5em}
    $^{\bbeta}$
    \vspace{8pt}   
  } \\
  {\bf
    Pradeep Dasigi
    \hspace{-0.5em}
    $^{\balpha}$
    \enskip
    Nathan Lambert 
    \hspace{-0.5em}
    $^{\balpha}$
    \enskip
    Kyle Richardson 
    \hspace{-0.5em}
    $^{\balpha}$
    \vspace{1.5pt}
  } \\
  {\bf
    Luke Zettlemoyer
    \hspace{-0.5em}
    $^{\bbeta}$
    Jesse Dodge
    \hspace{-0.5em}
    $^{\balpha}$
    \enskip
    Kyle Lo
    \hspace{-0.5em}
    $^{\balpha}$
    \enskip
    Luca Soldaini
    \hspace{-0.5em}
    $^{\balpha}$
    \vspace{8pt}   
  } \\
  {\bf
    Noah A. Smith
    \hspace{-0.5em}
    $^{\balpha}$$^{\bbeta}$
    \enskip
    Hannaneh Hajishirzi
    \hspace{-0.5em}
    $^{\balpha}$$^{\bbeta}$
    \vspace{8pt}
  } \\
  {
    $^{\balpha}$Allen Institute for Artificial Intelligence
   } \\
   {
    $^{\bbeta}$University of Washington \quad
    $^{\bgamma}$Yale University
  } \\
  {
    $^{\bdelta}$New York University \quad
    $^{\bmu}$Carnegie Mellon University
  }\vspace{8pt}\\
  \texttt{olmo@allenai.org}
}

\begin{document}
\maketitle
\begin{abstract}
Language models (LMs) have become ubiquitous in both NLP research and in commercial product offerings.
As their commercial importance has surged, the most powerful models have become closed off, gated behind proprietary interfaces, with important details of their training data, architectures, and development undisclosed. 
Given the importance of these details in scientifically studying these models, including their biases and potential risks, we believe it is essential for the research community to have access to powerful, truly open LMs.
To this end, we have built  \olmo, a competitive, truly \textbf{O}pen \textbf{L}anguage \textbf{Mo}del, to enable the scientific study of language models.
Unlike most prior efforts that have only released model weights and inference code, we release \olmo alongside open training data and training and evaluation code. We hope this release will empower
the open research community and inspire a new wave of innovation.
\end{abstract}

\section{Introduction}

Language models have been at the center of NLP technologies for many years \citep{rosenfeld2000two,Bengio2003ANP,Mikolov2013DistributedRO,Peters2018DeepCW,Brown2020LanguageMA}.
Recently, due to large-scale pretraining and human annotation for alignment, they have become commercially valuable \citep{OpenAI2023GPT4TR}.
However, as their commercial value has increased, the largest models have become gated behind proprietary interfaces, with important details left undisclosed.

We believe that full access to open language models for the research community is critical to the scientific study of these models, their strengths and weaknesses, and their biases and risks.
Accordingly, we introduce  \textbf{\olmo}, a powerful, truly open language model alongside open training data, training and evaluation code, intermediate model checkpoints, and training logs.  

Recent LM releases have varied in their degree of openness. For example, Mixtral 8x7B provided model weights and a brief report~\citep{jiang2024mixtral},
while LLaMA came with in-depth adaptation training instructions~\citep{touvron2023llama2}, and Mosaic Pretrained Transformer came with many details, including the dataset distribution, though not the data itself~\citep{MosaicML2023Introducing}.  Falcon's pretraining data was partially released~\citep{Falcon},
and the most open models---the Pythia suite~\citep{pmlr-v202-biderman23a} and BLOOM~\citep{workshop2022bloom}---released training code, model checkpoints, 
data, 
and more.  

\begin{table*}[t]
    \centering
    \begin{tabular}{c|cccc|ccccc}
        \toprule
        \textbf{Size} & \textbf{L} & \textbf{D} & \textbf{H} & \textbf{Tokens} & \textbf{Peak LR} & \textbf{Warmup} & \textbf{Weight Tying} & \textbf{Batch size} \\
        \hline
        1B & 16 & 2048 & 16 & 2T & \texttt{4.0E-4} & 2000 steps & yes & $\sim$4M \\
        7B & 32 & 4086 & 32 & 2.46T & \texttt{3.0E-4} & 5000 steps & no & $\sim$4M \\
        \bottomrule
    \end{tabular}
    \caption{\olmo model sizes, number of training tokens, and optimizer settings. In all runs, the optimizer was AdamW, with betas of $0.9$ and $0.95$, and an epsilon of \texttt{1.0E-5}.
    \textbf{L} is number of layers, \textbf{D} is hidden dimension, \textbf{H} is number of attention heads, \textbf{WD} is weight decay. 
    }
    \label{tab:model_size_and_optimizer}
\end{table*}

With \olmo, we release the whole framework from data to training to evaluation tools: multiple training checkpoints across multiple hardware types, training logs, and exact datasets used, with a permissive license.
We are not the only team to do this; recent work from LLM360 targets similar goals~\citep{liu2023llm360}.
\olmo~narrows the gap from their models to state-of-the-art capabilities of models like Llama 2. This project has benefited from lessons learned from all of these previous efforts with their varying degrees of openness, and we believe that a large, diverse population of open models is the best hope for scientific progress on understanding language models and engineering progress on improving their utility.



The \olmo framework encompasses the tools and resources required for building and researching language models. For training and modeling, it includes 
full model weights, training code, 
training logs, and inference code. The released model includes four variants of our language model at the 7B scale corresponding to different architectures, optimizers, and training hardware, and one model at the 1B scale, all trained on at least 2T tokens. We also release hundreds of intermediate checkpoints available as revisions on HuggingFace.
For dataset building and analysis, the full training data used for these models is openly available (Dolma; \citealp{dolma}), including code that produces the training data, and tools for analyzing pretraining data  \citep{wimbd}. 
For evaluation, we build on Catwalk~\citep{groeneveld2023catwalk} for downstream evaluation and Paloma~\citep{magnusson2023paloma} for perplexity-based evaluation.
For adaptation, we use Open Instruct~\citep{ivison2023camels,wang2023far} to train with instruction and feedback data. 
Finally, all code and weights are released under the Apache 2.0 License.\footnote{\url{https://allenai.org/olmo}}

With this release, we hope to catalyze research into as-yet poorly understood aspects of these models, for example, the relationship between pretraining data and model capabilities, the impact of design and hyperparameter choices, and various optimization methods and their impact on model training. In addition, we report on the lessons learned and important details necessary to successfully train language models at this scale.




\section{\olmo Framework}
This section describes the \olmo~framework, consisting of the \olmo models (Section~\ref{sec:olmo-arch}), our pre-training dataset, \dolma (Section~\ref{sec:pretraining-data}), and our evaluation framework (Section~\ref{sec:evaluation}). 

\subsection{\olmo~Model and Architecture}\label{sec:olmo-arch}

We adopt a decoder-only transformer architecture based on \cite{vaswani2017attention},
and deliver 1B and 7B variants as described in Table \ref{tab:model_size_and_optimizer}.
Our specific architecture includes several improvements over the vanilla transformer from \cite{vaswani2017attention} following other recent large language models 
like PaLM \citep{chowdhery2022palm}, the LLaMA family \citep{touvron2023llama1, touvron2023llama2}, OpenLM \citep{open_lm}, and Falcon \citep{Falcon}.
See Table~\ref{tab:llm_arch_adamw} in Appendix~\ref{sec:training_details}  for a comprehensive comparison of our 7B architecture to the similarly-sized models from these other families.

We generally select hyperparameters by optimizing  for training throughput on our hardware while minimizing the risk of loss spikes and slow divergence. We ablate choices through our in-loop evaluation setting, given available computational sources (Section~\ref{sec:ablation}).  
Our main changes over the vanilla transformer architecture can be summarized as follows:

\begin{enumerate}[itemsep=1mm,leftmargin=4.5mm]
    \item \textbf{No biases.} Following LLaMA, PaLM, and others, we exclude all bias terms from our architecture in order to improve training stability.

    \item \textbf{Non-parametric layer norm.} We use the non-parametric formulation of layer norm \citep{Ba2016LayerNorm} in which there is no affine transformation within the norm, i.e., no ``adaptive gain" (or bias). We believe this was the safest option and it was also the fastest compared to the other variants we considered: parametric layer norm and RMSNorm \citep{RMSNorm}.

    \item \textbf{SwiGLU activation function.} Like LLaMA, PaLM, and others we use the SwiGLU activation function \citep{Shazeer2020GLUVI} instead of ReLU, and following LLaMA the activation hidden size is approximately $\frac{8}{3}d$, but increased to the closest multiple of 128 (e.g. 11,008 for our 7B model) to improve throughput.\footnote{Since SwiGLU is a ``gated" activation function, the output is half the size of the input. So technically our inputs to SwiGLU have a dimensionality of 2 $\times$ 11,008 = 22,016 for our 7B model.}

    \item \textbf{Rotary positional embeddings (RoPE).} Like LLaMA, PaLM, and others we replace absolute positional embeddings with rotary positional embeddings (RoPE; \citealp{Su2021RoFormerET}).

    \item \textbf{Vocabulary.} We use a modified version of the BPE-based tokenizer from GPT-NeoX-20B \citep{gpt-neox-20b} with additional tokens for masking personal identifiable information (PII). The final vocabulary size is 50,280.
    However, to maximize training throughput we increase the size of the corresponding embedding matrix in our model to 50,304 
    to be
    a multiple of 128.
\end{enumerate}


\subsection{Pretraining Data: \dolma} \label{sec:pretraining-data}

Despite progress in access to model parameters, pretraining datasets are still not as open. 
Pretraining data are often not released alongside open models (let alone closed models) and documentation about such data is often lacking in detail that would be needed to reproduce or fully understand the work. 
This has made it difficult to support certain threads of language model research, such as understanding how training data impacts model capabilities and limitations.
To facilitate open research on language model pretraining, we built and released our pretraining dataset, \dolma---a diverse, multi-source corpus containing trillions of tokens across billions of documents acquired from different data sources that are (1) commonly seen in large-scale language model pretraining and (2) accessible to the general public \citep{dolma}. 
Table~\ref{tab:statistics} provides a high-level overview of the amount of data from each source.

\dolma is built using a pipeline of 
(1) language filtering,
(2) quality filtering,
(3) content filtering,
(4) deduplication,
(5) multi-source mixing, and 
(6) tokenization.
We refer the reader to the \dolma report~\citep{dolma} for more details about its design principles, details about its construction, and a more detailed summary of its contents.
The report provides additional analyses and experimental results from training language models on intermediate states of \dolma to share what we learned about important data curation practices, including the role of content or quality filters, deduplication, and mixing data from multiple sources.
We keep documents from each source separate, both during curation as well as in the final release.
We open-sourced our high-performance data curation tools;
this toolkit can be used to further experiment on Dolma, reproduce our work, and enable fast and easy curation of pretraining corpora.
Finally, we also open-sourced our WIMBD tool~\citep{wimbd} to help with dataset analysis. 


\begin{table}[t]
\centering
\small
\begin{tabular}{l@{}c@{}c@{}c@{}c@{}}
\toprule
\textbf{ Source } & \textbf{ Type } & {\begin{tabular}[c]{@{}c@{}}\textbf{ UTF-8}\\\textbf{bytes}\\\textit{(GB)}\vspace{.4em}\end{tabular}} & {\begin{tabular}[c]{@{}c@{}}\textbf{ Docs }\\\textit{(millions) }\vspace{.4em}\end{tabular}} & {\begin{tabular}[c]{@{}c@{}}\textbf{Tokens}\\\textit{(billions) }\vspace{.4em}\end{tabular}}  \\
\midrule
Common Crawl & web pages & 9,812 & 3,734 & 2,180  \\
GitHub & code & 1,043 & 210 & 342  \\
Reddit & social media & 339 & 377 & 80  \\
Semantic Scholar & papers & 268 & 38.8 & 57  \\
Project Gutenberg & books & 20.4 & 0.056 & 5.2 \\
Wikipedia & encyclopedic & 16.2 & 6.2 & 3.7 \\
\midrule
\multicolumn{2}{c}{\textbf{Total}} & \textbf{11,519} & \textbf{4,367} & \textbf{2,668}  \\
\bottomrule
\end{tabular}
\caption{Composition of \dolma. Tokens counts are based on the GPT-NeoX tokenizer.}
\label{tab:statistics}
\end{table}

\subsection{Adaptation}
\label{sec:adaptation}

Pretrained models are not always used as-is, but rather further finetuned to improve their performance, safety, and usability. Often models are first trained to follow instructions~\citep{mishra-etal-2022-cross, wei2022finetuned, sanh2022multitask}, and then further trained on human preferences~\citep{NEURIPS2022_b1efde53} to improve the quality of their generations. We showcase the efficacy of using \olmo as a base model for further fine-tuning by training \olmo to be a general chat assistant following the \tulu data and training setup~\citep{ivison2023camels}. This involves first performing instruction finetuning with a mixture of distilled and human-written instruction data and then further aligning the model with distilled preference data using Direct Preference Optimization (DPO)~\citep{rafailov2023direct}.

\subsection{Evaluation}
\label{sec:evaluation}
We perform base model evaluation at two stages: \emph{online} evaluation to make decisions for model design and \emph{offline} evaluation to evaluate model checkpoints. 
For the offline stage, we use the Catwalk framework \citep{groeneveld2023catwalk}, a publicly available evaluation tool with access to a wide range of datasets and task formats, 
to perform downstream evaluation as well as intrinsic language modeling evaluation on the perplexity benchmark Paloma \citep{magnusson2023paloma}.

For both downstream and perplexity evaluation, we use our fixed evaluation pipeline
to compare results against publicly available models. We also report a separate evaluation of our adapted model.  

\paragraph{In-Loop Training Ablations}
\label{sec:ablation}

Throughout model training, we perform downstream evaluations to make decisions around model architecture, initialization, optimizers, learning rate schedule, and data mixtures. We call this our \emph{online} evaluation as it runs in-loop every 1000 training steps (or $\sim$4B training tokens) and provides an early and continuous signal on the quality of the model being trained. These evaluations rely on many of the core tasks and experiment settings used for our \emph{offline} evaluation detailed in Section~\ref{sec:downstream_evaluation}, which also mirrors the task and evaluation structure of the EleutherAI eval harness~\citep{eval-harness}.

\paragraph{Downstream Evaluation}
\label{sec:evaluation-downstream}
Following much previous work \citep[][\emph{inter alia}]{Brown2020LanguageMA,gpt-neox-20b,touvron2023llama1,touvron2023llama2}, we report zero-shot performance on a set of downstream tasks. Our evaluation suite consists of 8 core tasks corresponding closely to the commonsense reasoning task set reported by \citet{touvron2023llama1} and \citet{touvron2023llama2} (see Table~\ref{table:results-end-task} for a list of tasks). Given the scale of the models being evaluated, such tasks were selected at the beginning of model development due to their naturalness (e.g., all can formulated as text completion scoring tasks) and ability to provide meaningful signals throughout training (see Figure~\ref{fig:core-tasks-progression}).

\paragraph{Intrinsic Language Modeling Evaluation}

To measure how OLMo fits distributions of language beyond held-out training data, we use Paloma \citep{magnusson2023paloma}, a new perplexity benchmark that includes 585 different domains of text. Domains range from nytimes.com to r/depression on Reddit and are drawn from 18 separate data sources, such as C4 \citep{raffel2020exploring}, in stratified samples. This allows for more equal inclusion of text domains that are under-represented in their source corpora. 

We aim not just to compare \olmo against other models for best performance, but also to demonstrate how it enables fuller and more controlled scientific evaluations. \model is the largest LM with explicit decontamination for perplexity evaluation. Following the approach described in Paloma, we remove any pretraining document with paragraphs leaked from Paloma evaluation data. Without decontamination, other models risk underestimating perplexity (i.e., overestimating the model's out-of-sample fit).  We also release intermediate checkpoints, allowing richer comparisons with two other models that release checkpoints, Pythia-6.9B \citep{pmlr-v202-biderman23a} and RPJ-INCITE-7B \citep{together2023redpajama} (see Figure~\ref{fig:paloma_bpb_over_tokens_seen_per_source_clean}). 

\paragraph{Adaptation Evaluation} We also evaluate \olmo after instruction fine-tuning and DPO training using the \tulu evaluation suite proposed in \citet{wang2023far, ivison2023camels}.
We focus on evaluations around model chat capabilities and safety in order to showcase the efficacy of using \olmo as a base for further fine-tuning.

\section{Training \olmo}

This section describes our pretraining setup, including our distributed training framework (Section~\ref{sec:training-distributed}), optimizer  (Section~\ref{sec:training-optimizer}), data preparation (Section~\ref{sec:training-data}), and hardware (Section~\ref{sec:hardware}).

\subsection{Distributed Training Framework}
\label{sec:training-distributed}

We train our models using the \emph{ZeRO} optimizer strategy \citep{Rajbhandari2019ZeRO} via PyTorch's \texttt{FSDP} framework \citep{Zhao2023PyTorchFSDP}, which reduces memory consumption by sharding the model weights and their corresponding optimizer state across GPUs.
At the 7B scale, this enables training with a micro-batch size of 4096 tokens per GPU on our hardware (see Section \ref{sec:hardware}).
For \olmo-1B and -7B models, we use a constant global batch size of approximately 4M tokens (2048 instances, each with a sequence length of 2048 tokens).

To improve throughput, we employ mixed-precision training \citep{Micikevicius2017MixedPT} through \texttt{FSDP}'s built-in settings and PyTorch's \texttt{amp} module.
The latter ensures that certain operations like the softmax always run in full precision to improve stability, while all other operations run in half-precision with the \texttt{bfloat16} format. 
Under our specific settings, the sharded model weights and optimizer state local to each GPU are kept in full precision.
The weights within each transformer block are only cast to \texttt{bfloat16} when the full-sized parameters are materialized on each GPU during the forward and backward passes.
Gradients are reduced across GPUs in full precision.

\subsection{Optimizer}
\label{sec:training-optimizer}

We use the AdamW optimizer \citep{loshchilov2018decoupled} with the hyperparameters shown in Table \ref{tab:model_size_and_optimizer}.
For all model sizes, we warm up the learning rate over 5000 steps ($\sim$21B tokens) and then decay it linearly from there down to a tenth of the peak learning rate over the remainder of training.
After the warm-up period, we clip gradients such that the total $l^2$-norm of the parameter gradients\footnote{During gradient clipping all of the model's parameters are treated as a single big vector (as if all parameters were flattened and concatenated together), and we take the $\ell_2$-norm over the corresponding single gradient vector. This is the standard way to clip gradients in PyTorch.} does not exceed $1.0$. Table \ref{tab:llm_arch_adamw} gives a comparison of our optimizer settings at the 7B scale to those of other recent LMs that also used AdamW.

\subsection{Data}
\label{sec:training-data}

We built our training dataset out of a 2T-token sample from our open dataset, Dolma \citep{dolma}, which we describe in Section \ref{sec:pretraining-data}.
The tokens from every document are concatenated together after appending a special \texttt{EOS} token to the end of each document, and then we group consecutive chunks of 2048 tokens to form training instances.
The training instances are shuffled in the exact same way for each training run.
The data order and exact composition of each training batch can be reconstructed from the artifacts we release.

All of our released models have been trained to at least 2T tokens (a single epoch over our training data), and some have been trained beyond that by starting a second epoch over the data with a different shuffling order. The impact of repeating this small amount of data should be negligible according to prior work~\citep{muennighoff2023scaling}.

\subsection{Hardware} \label{sec:hardware}
In order to verify that our codebase could be used on both NVIDIA and AMD GPUs without any loss in performance, we trained models on two different clusters:
\begin{itemize}[noitemsep,leftmargin=4.5mm]
    \item \textbf{LUMI:} Provided by the LUMI supercomputer,\footnote{\url{https://www.lumi-supercomputer.eu}} we used up to 256 nodes on this cluster, 
where each node consists of 4x AMD MI250X GPUs with 128GB of memory\footnote{The MI250X is a dual-chip module, meaning in practice that each physical device consists of two logical devices, so each node has 8 logical GPU devices with 64GB of memory each.} and 800Gbps of interconnect.
    \item \textbf{MosaicML:} Provided by MosaicML\footnote{\url{https://www.mosaicml.com}} (Databricks), we used 27 nodes on this cluster, where each node consists of 8x NVIDIA A100 GPUs with 40GB of memory and 800Gbps interconnect.
\end{itemize}
Despite minor differences in batch size to optimize for training throughput, both runs resulted in nearly identical performance on our evaluation suite by 2T tokens.


\section{Results}
\label{sec:results}
The checkpoint used for evaluating \model is trained until 2.46T
tokens on the Dolma~\citep{dolma} dataset with a linear learning rate decay schedule mentioned in Section~\ref{sec:training-optimizer}. In our experiments, we find that tuning this checkpoint further on the Dolma dataset for 1000 steps with the learning rate linearly decayed to 0 boosts model performance on perplexity and end-task evaluation suites described in Section~\ref{sec:evaluation}.  We compare \olmo with other publicly available models including LLaMA-7B \citep{touvron2023llama1}, Llama-2-7B \citep{touvron2023llama2}, MPT-7B \citep{MosaicML2023Introducing}, Pythia-6.9B \citep{pmlr-v202-biderman23a}, Falcon-7B \citep{Falcon} and RPJ-INCITE-7B \citep{together2023redpajama}.



\begin{table*}[t]
\centering 
\begin{tabular}{@{}l|cccccccc|c
}
\toprule
{\bf Models} & {\renewcommand{\arraystretch}{1}\begin{tabular}[c]{@{}c@{}}{arc}\\{challenge}\\\end{tabular}} & {\renewcommand{\arraystretch}{1}\begin{tabular}[c]{@{}c@{}}{arc}\\{easy}\\\end{tabular}} &  boolq & {\renewcommand{\arraystretch}{1}\begin{tabular}[c]{@{}c@{}}{hella-}\\{swag}\\\end{tabular}} & {\renewcommand{\arraystretch}{1}\begin{tabular}[c]{@{}c@{}}{open}\\{bookqa}\\\end{tabular}} & piqa & sciq & {\renewcommand{\arraystretch}{1}\begin{tabular}[c]{@{}c@{}}{wino-}\\{grande}\\\end{tabular}} & avg. \\
\hline
\textbf{StableLM 1.6B}      & 43.8 & 63.7 & 76.6 & 68.2 & 45.8 & 74.0 & 94.7 & 64.9 & 66.5 \\
\textbf{Pythia 1B}      & 33.1 & 50.2 & 61.8 & 44.7 & 37.8 & 69.1 & 86.0 & 53.3 & 54.5 \\
\textbf{TinyLlama 1.1B}      & 34.8 & 53.2 & 64.6 & 58.7 & 43.6 & 71.1 & 90.5 & 58.9 & 59.4 \\
\rowcolor{olmocolor!50}[\dimexpr\tabcolsep+0.1pt\relax] \textbf{\modelsmall}         & 34.5 & 58.1 & 60.7 & 62.5 & 46.4 & 73.7 & 88.1 & 58.9 & 60.4 \\
\hline
\textbf{Falcon-7B}      & 47.5 & 70.4 & 74.6 & 75.9 & 53.0 & 78.5 & 93.9 & 68.9 & 70.3 \\
\textbf{LLaMA 7B}       & 44.5 & 67.9 & 75.4 & 76.2 & 51.2 & 77.2 & 93.9 & 70.5 & 69.6 \\
\textbf{Llama 2 7B}      & 48.5 & 69.5 & 80.2 & 76.8 & 48.4 & 76.7 & 94.5 & 69.4 & 70.5 \\
\textbf{MPT-7B}         & 46.5 & 70.5 & 74.2 & 77.6 & 48.6 & 77.3 & 93.7 & 69.9 & 69.8 \\
\textbf{Pythia 6.9B}      & 44.1 & 61.9 & 61.1 & 63.8 & 45.0 & 75.1 & 91.1 & 62.0 & 63.0 \\
\textbf{RPJ-INCITE-7B}  & 42.8 & 68.4 & 68.6 & 70.3 & 49.4 & 76.0 & 92.9 & 64.7 & 66.6 \\
\rowcolor{olmocolor!50}[\dimexpr\tabcolsep+0.1pt\relax] \textbf{\model}         & 48.5 & 65.4 & 73.4 & 76.4 & 50.4 & 78.4 & 93.8 & 67.9 & 69.3 \\

\bottomrule
\end{tabular}
\caption{Zero-shot evaluation of \modelsmall and \model, with other publicly available comparable model checkpoints on 8 core tasks from the downstream evaluation suite described in Section~\ref{sec:evaluation-downstream}. For \model, we report results for the 2.46T token checkpoint.}
\label{table:results-end-task}
\end{table*}

\subsection{Downstream evaluation} 
\label{sec:downstream_evaluation}
\paragraph{Setup} Our core \textbf{downstream evaluation suite}  (see Table~\ref{table:results-end-task}) consists of: arc (both \texttt{arc\_easy} and \texttt{arc\_challenge}) \citep{clark2018think}, \texttt{boolq} \citep{clark2019boolq}, \texttt{openbookqa} \citep{mihaylov2018can}, \texttt{sciq} \citep{welbl2017crowdsourcing}, \texttt{hellaswag} \citep{zellers2019hellaswag}, \texttt{piqa} \citep{bisk2020piqa}, 
and \texttt{winogrande} \citep{sakaguchi2021winogrande}. In Appendix~\ref{sec:add_eval}, we also report results on an additional set of auxiliary tasks outside of our core evaluation set that we found to have less stable performance trends (see Figure~\ref{fig:additional-task-acc-progression}). 

In all cases, we perform zero-shot evaluation using the rank classification approach popularized by \citet{Brown2020LanguageMA}. Under this approach, candidate text completions (e.g., different multiple-choice options) are ranked by likelihood (usually normalized by some normalization factor), and prediction accuracy is reported. While Catwalk implements several common likelihood normalization strategies, including normalizing by number of tokens (per-token normalization; \citealp{Brown2020LanguageMA,liang2022holistic}), by number of characters (per-character normalization; \citealp{eval-harness}), as well as incorporating an answer's unconditional likelihood \citep{Brown2020LanguageMA}, we selected the normalization strategies for each dataset separately. Specifically, we used unconditional normalization for \texttt{arc} and \texttt{openbookqa}, per-token normalization for \texttt{hellaswag}, \texttt{piqa}, and \texttt{winogrande} and no normalization for \texttt{boolq}, 
and \texttt{sciq} (i.e., tasks formulated as single token prediction tasks).

\paragraph{Results} Table~\ref{table:results-end-task} summarizes the result of zero-shot evaluation of \olmo and compares against other publicly available models of comparable size. We report results on 8 core tasks from our evaluation suite described in Section~\ref{sec:evaluation-downstream}. On aggregate, \model is competitive against all the comparable models. 
We include the comparison to StableLM 1.6B
, but note that it is significantly larger, and was trained on unknown data.

\begin{figure*}[!t]
\centering
\makebox[\columnwidth][c]{
\includegraphics[width=0.80\textwidth]{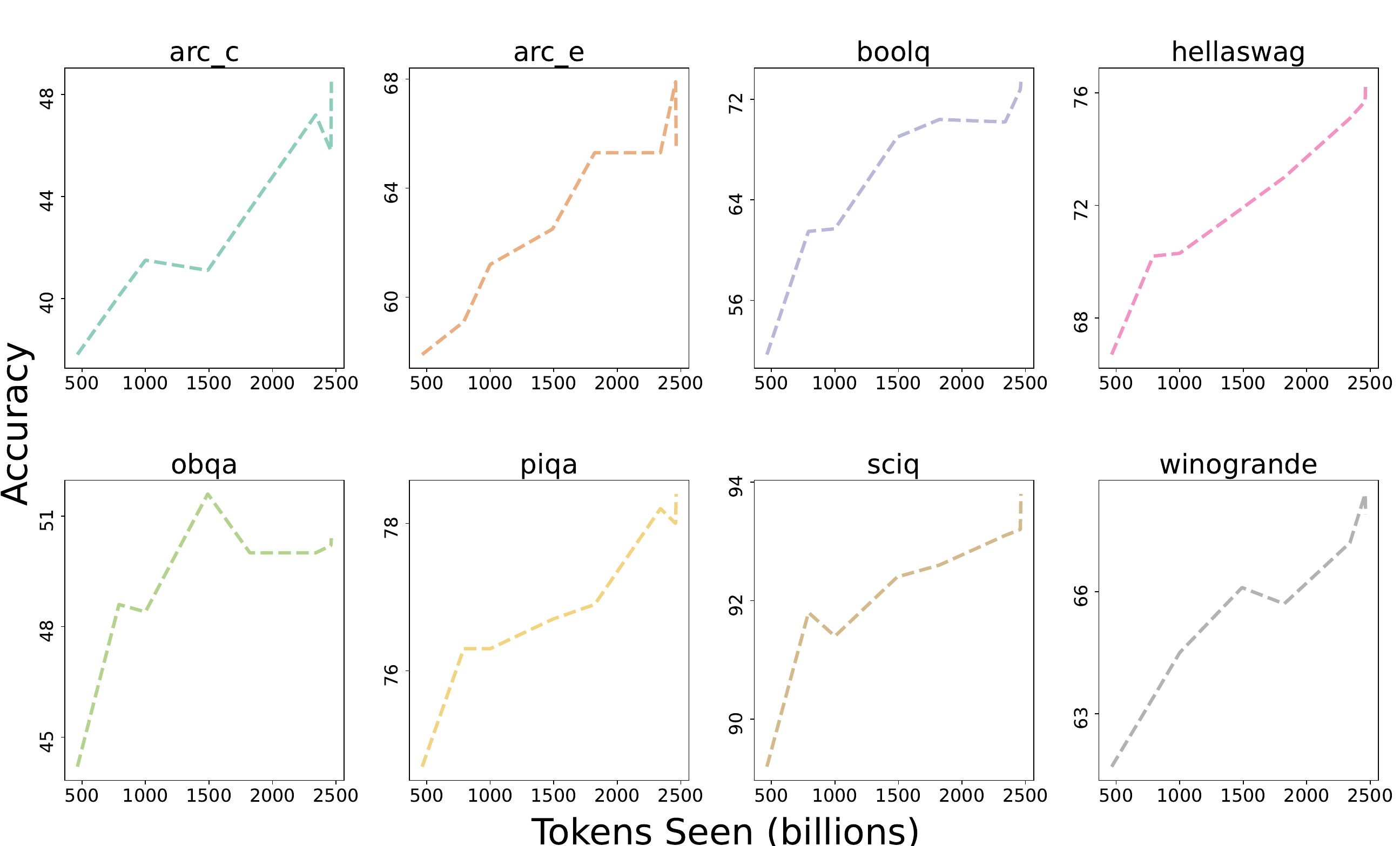}
}
\caption{Accuracy score progression of \model on 8 core end-tasks score from Catwalk evaluation suite described in Section~\ref{sec:evaluation-downstream}. We can see the benefit of decaying LR to 0 in the final 1000 steps of training on most tasks.}
\label{fig:core-tasks-progression}
\end{figure*}

In Figure~\ref{fig:core-tasks-progression} we plot the accuracy score progression of 8 core end-tasks. All tasks, except OBQA, show an upward trend in accuracy numbers as \model is trained on more tokens. A sharp upward tick in accuracy of many tasks between the last and the second to last step shows us the benefit of linearly reducing the LR to 0 over the final 1000 training steps. See Table~\ref{table:additional-end-task} in Appendix~\ref{sec:add_eval} for additional evaluation results and discussion.

\subsection{Intrinsic language modeling evaluation} 
\paragraph{Setup}
For intrinsic evaluations, Paloma proposes a range of analyses, from inspection of performance in each domain separately to more summarized results over combinations of domains. We report results at two levels of granularity: the aggregate performance over 11 of the 18 sources in Paloma as in \cite{magnusson2023paloma}, as well as more fine-grained results over each of these sources individually. 
This particular subset of 11 sources from Paloma excludes sources that are not publicly available, involve fringe or toxic text, or consist of code data not supported by Paloma's decontamination approach. 
This leaves C4 \citep{raffel2020exploring}, mC4-en \citep{chung2023unimaxfa}, Wikitext 103 \citep{merity2016pointersm}, Penn Treebank \citep{marcusptb,nunesptb}, RedPajama \citep{together2023redpajama}, Falcon-RefinedWeb \citep{penedo2023therd}, Dolma \citep{dolma}, M2D2 S2ORC \citep{reid-etal-2022-m2d2}, M2D2 Wikipedia \citep{reid-etal-2022-m2d2}, C4 100 domains \citep{chronopoulou-etal-2022-efficient}, and Dolma 100 Subreddits \citep{dolma}. To allow for a fair comparison between models with different vocabularies, we report bits per byte as defined by \citet{gao2020pile} over the test sets of these sources.

\begin{figure*}
    \centering
    \includegraphics[width=0.85\textwidth]{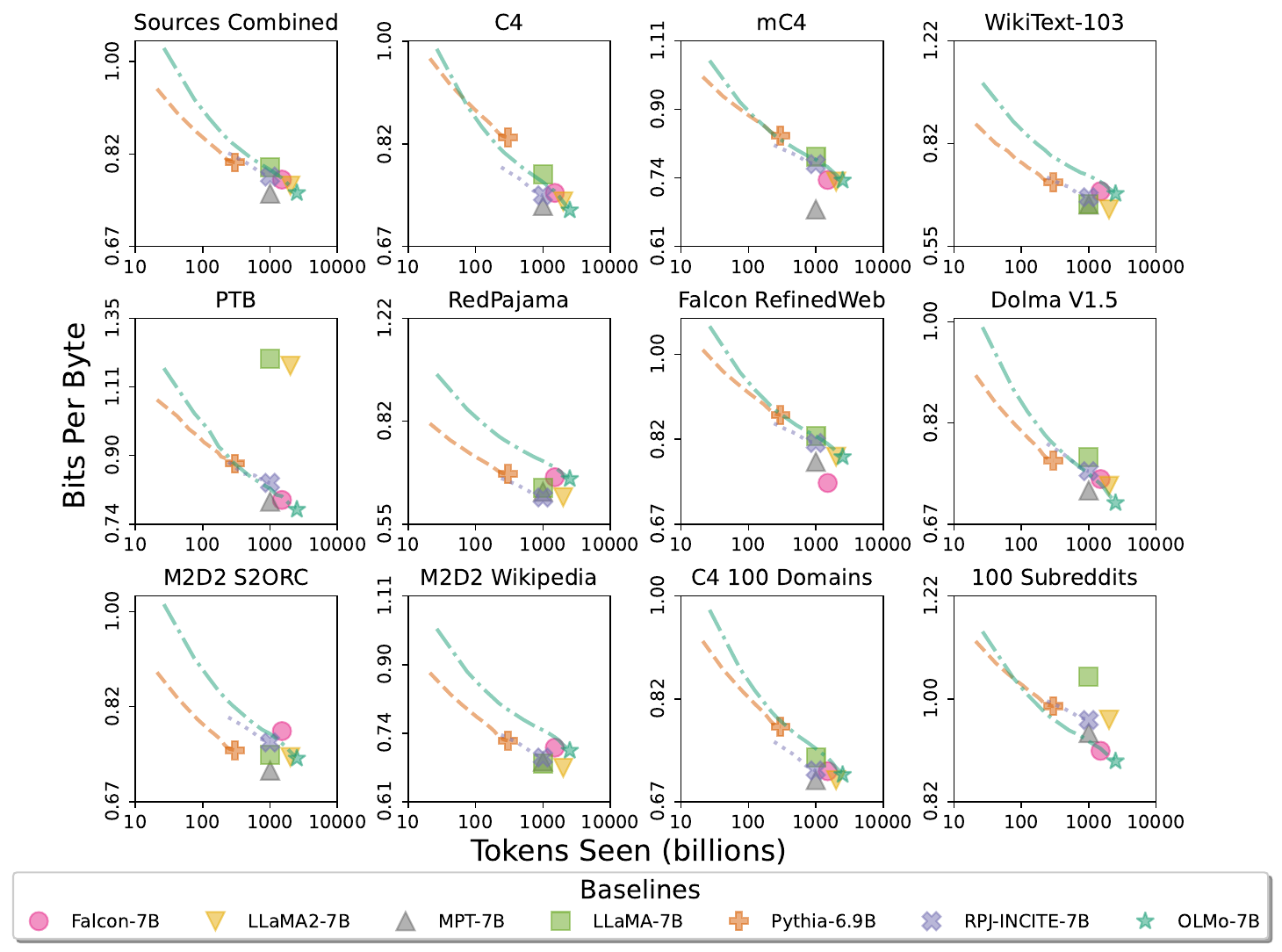}
    \caption{
    Bits per byte on 11 evaluation data sources from Paloma and their combination \citep{magnusson2023paloma}, decontaminated from \olmo's pretraining data. While models follow a general data scaling trend, sample efficiency is most favorable on in-distribution data. For example, \model overtakes all other models on C4, perhaps from having 88.8\% Common Crawl pretraining data.
    }
    \label{fig:paloma_bpb_over_tokens_seen_per_source_clean}
\end{figure*}

\paragraph{Results} In the \textit{Sources Combined} subplot of Figure~\ref{fig:paloma_bpb_over_tokens_seen_per_source_clean}, we show the performance of \model against 6 comparably-sized language models on the combination of 11 data sources from Paloma. Overall we find \olmo to have a competitive fit, especially given its training data was explicitly decontaminated against Paloma. As seen through the comparison of final models (see shapes) as well intermediate checkpoints (see dashed lines),  the \olmo results follow similar scaling trends of other models. Note that the performance of intermediate checkpoints is influenced by where that checkpoint occurs in the learning rate schedule. So models trained for fewer steps will tend to have steeper training curves without necessarily being more sample efficient if training duration were fixed across all models. MPT-7B, nevertheless, stands out as improving ahead of the other models in this subplot. This could be due to a number of factors, including pretraining data composition and its match to the domains in Paloma (e.g., MPT trains on 27\%  non-Common Crawl data rather than 18\% for LLaMA, 12.2\% for RedPajama, and 11.2\% for \olmo) as well as various data preprocessing decisions (e.g., MPT's use of semantic deduplication by \citealp{abbas2023semdedup}, on C4).





The remaining subplots in Figure~\ref{fig:paloma_bpb_over_tokens_seen_per_source_clean} provide more fine-grained analysis by reporting bits per byte separately for each of the 11 data sources that are combined in the aggregated Paloma metric. From this we see greater variation in sample efficiency, largely driven by the similarity of training and evaluation distributions. Notably, \model fares well on evaluations predominated by Common Crawl, such as C4, though different ways of postprocessing Common Crawl are best fit by models trained with that specific data, such as Falcon-7B on Falcon RefinedWeb.  Meanwhile, \model is less sample efficient compared to other models on sources less related to scraped web text, such as WikiText-103, M2D2 S2ORC, and M2D2 Wikipedia. The RedPajama evaluation shows a similar pattern, perhaps as only 2 of its 7 domains are from Common Crawl, and Paloma weights domains within each source equally. Since heterogeneous data from curated sources like Wikipedia and ArXiv papers is 
scarcer
than scraped web text, maintaining sample efficiency for fit to these distributions of language will be challenging as pretraining corpora are scaled.

\subsection{Adaptation Evaluation}

\begin{table}[t]
\setlength{\tabcolsep}{1pt}
\adjustbox{max width=\columnwidth}{
\centering
\begin{tabular}{@{}l|cccc
}
\toprule
\textbf{Model} & \textbf{MMLU}  & \textbf{AlpacaEval} & \textbf{ToxiGen} & \textbf{TruthfulQA}    \\
      & \textbf{0-shot} $\uparrow$ & \textbf{\%win} $\uparrow$ & \textbf{\% Toxic} $\downarrow$ & \textbf{\%Info+True} $\uparrow$ \\
\hline
 \textbf{\olmo (base)}        & 28.3 & - & 81.4 & 31.6 \\ \hline
\textbf{MPT Chat} & 33.8 & 46.8 & 0.1 & 42.7 \\
\textbf{Falcon Instruct} & 25.2 & 14.0 & 70.7 & 27.2 \\
\textbf{RPJ-INCITE Chat} & 27.0 & 38.0 & 46.4 & 53.0 \\
\textbf{Llama-2-Chat} & 46.8 & 87.3 & 0.0 & 26.3 \\ \hline
\textbf{\tulu 2}      & 50.4 & 73.9 & 7.0 & 51.7 \\
\textbf{\tulu 2+DPO}      & 50.7 & 85.1 & 0.5 & -\footnotemark \\
\rowcolor{olmocolor!50}[\dimexpr\tabcolsep+0.1pt\relax] \textbf{OLMo+SFT}        & 47.3 & 57.0 & 14.4 & 41.2 \\
\rowcolor{olmocolor!50}[\dimexpr\tabcolsep+0.1pt\relax] \textbf{OLMo+SFT+DPO}        & 46.2 & 69.3 & 1.7 & 52.0 \\
\bottomrule
\end{tabular}
}
\caption{Evaluation of various instruction-tuned 7B models, including \model and before and after adaptation training. Lower is better for ToxiGen and higher is better for other metrics. We provide a detailed description of models and metrics in Appendix.~\ref{app:adaptation-evals-description}.}
\label{table:results-adaptation}
\end{table}

\paragraph{Setup} We evaluate \model before adaptation, and after both the supervised fine-tuning and DPO training stage, focusing on the safety and chat evaluations used by \citet{wang2023far}. We additionally compare to officially released instruction-tuned variants of the models from Table~\ref{table:results-end-task}. We finally also compare to \tulu 2 models to compare against models trained using the same post-training data mixes and procedures.

\footnotetext{Following \citet{ivison2023camels}, we do not report \tulu 2 TruthfulQA scores due to test set contamination.}

\paragraph{Results} We find that instruction tuning considerably improves the performance and safety of \model, increasing MMLU performance by a wide margin and improving ToxiGen and TruthfulQA scores - especially after DPO training. Additionally, we find that \model outperforms most other chat variants after both initial instruction tuning (OLMo+SFT) and additional preference alignment (OLMo+SFT+DPO), highlighting both the strength of \model as a base model and the strength of the \tulu mix used to perform adaptation training. However, we find there is still a gap with \tulu 2, which is trained by applying the \tulu mix on Llama 2. This gap may be due to test set contamination in Llama 2\footnote{\citet{touvron2023llama2} report that Llama 2 was pretrained on data contaminated with MMLU test data.} and because the \tulu mix was primarily designed for Llama models. Overall, we see that \model greatly benefits from additional tuning and serves as a strong base model for downstream applications.

\section{Artifacts Released}
\label{sec:release}
By sharing artifacts from all pipeline stages, we aim to encourage open research and reduce duplicated, often costly efforts, by academics and practitioners. We release the following:

\begin{itemize}[noitemsep,leftmargin=4.5mm,beginpenalty=10000]
    \item \textbf{Pretraining (\S\ref{sec:olmo-arch})} 
    \begin{enumerate}
        \item The training and modeling code.
        \item The trained model weights for the 7B model, 7B-twin-2T, and the 1B model. For all the models, we release not only the final model weights but also 500+ intermediate checkpoints at intervals of 1000 steps.
        \item The complete set of metrics logged to Weights \& Biases during training.
    \end{enumerate}
    \item \textbf{Data (\S\ref{sec:pretraining-data})} 
    \begin{enumerate}
        \item Our full pretraining corpus Dolma~\citep{dolma}.
        \item Tools to support reproduction of full training data order as well as inspection of which training data was seen at each step during training.
        \item Tools for recreating our training data~\citep{dolma} and performing dataset analysis~\citep{wimbd}.
    \end{enumerate}
    \item \textbf{Adaptation (\S\ref{sec:adaptation})}
    \begin{enumerate}
        \item The training code and data for adaptation.
        \item The model weights for OLMo+SFT and OLMo+SFT+DPO.
    \end{enumerate}
    \item \textbf{Evaluation (\S\ref{sec:evaluation})}
    \begin{enumerate}
        \item The code and data in our evaluation framework Catwalk~\citep{groeneveld2023catwalk} for offline evaluation on both downstream tasks and intrinsic language modeling~\citep{magnusson2023paloma}.
        \item The evaluation suite~\citep{wang2023far,ivison2023camels} for adapted models.
    \end{enumerate}
\end{itemize}

\section{Conclusion and Future Work}

This paper presents our first release of  \olmo, a state-of-the-art, truly open language model and its framework to build and study the science of language modeling.
Unlike most prior efforts that have only released model weights and inference code, we release \olmo and the whole framework, including training data, training and evaluation code, and detailed metrics collected during the training runs. 
Additionally, we released adapted models, as well as all of our model adaptation code and data.

We intend to continuously support and extend \olmo and its framework, and continue to push the boundaries of open LMs to empower the open research community.
Since the original release of \olmo described here, we improved our data and training setup to significantly improve results. For example, MMLU scores have improved by 24 points to 52\%.\footnote{
\href{https://blog.allenai.org/olmo-1-7-7b-a-24-point-improvement-on-mmlu-92b43f7d269d}{\nolinkurl{https://medium.com/p/92b43f7d269d}}}
We look forward to bringing different model sizes, modalities, datasets, safety measures, and evaluations into the OLMo family.
We hope this and future releases will empower and strengthen the open research community and inspire a new wave of innovation.

\section*{Limitations}
We recognize building a large language model has many limitations. In fact, each step of the process of creating a language model, from the data to training to adaptation to evaluation each have their own limitations, and so we've added sections for each below. Of course we recognize that AI systems today can have broad societal reach, and therefore there are significant limitations beyond what we are able to fit into this section.

\paragraph{Data}
Our work focuses on pretraining data in English. We hope that our open framework enables the development of future models in more languages as well as multilingual models.
The data that models are trained on is what gives models their capabilities, and at the scale of training a large language model we recognize that the data likely contains problematic content like toxic language, personal information, and copyrighted text.
We mitigated this to the best of our ability but recognize there are no perfect approaches today that can completely remove such content.

\paragraph{Training}
Training a large language model is currently a challenging endeavor which is missing significant support from the open source community. With our limited page count we did not provide extensive training logs documenting, for example, training runs that diverged or failed to learn.

\paragraph{Adaptation}
Our pretrained models face the same issues as existing pretrained LLMs, such as bias, toxicity and, hallucinations. Our adapted models are better at avoiding these generations, but they are not perfect.
Additionally, we note that we largely adopt an existing data mixture designed for a different model family (\tulu, designed for Llama models), and \olmo may require different data mixing to adjust for its unique strengths and weaknesses. The \tulu mix itself also relies on data distilled from a variety of models, and we hope to reduce our reliance on such data in the future.

\paragraph{Evaluation}
While we've included comparisons on a variety of datasets to other current language models, many of the downstream tasks are not actually representative of how users interact with language models (i.e., as a chatbot). In addition, language model evaluations are currently very noisy; we aimed to include only evaluations on datasets that provided some signal as to which model performs best, but recognize that there is no perfect automatic evaluation, and thus comparisons should be taken with a grain of salt. 

\section*{Ethics Statement}
Through this work, we take the position that increased openness of language models is essential for scientific understanding of their abilities and limitations and for broad participation in the continued development of such models.  
Training on open data further enhances these benefits. In addition, our open release enables practitioners to take our models and build on them instead of having to train their own from scratch, in which case they would be repeating our work while consuming more resources and leading to an increased environmental impact. 
Of course, openness is not without risk; the possibility remains that these models will be used in unintended ways that cause harm.  
We believe that research and development efforts to understand and mitigate those potential harms will also be accelerated by the openness of the models, allowing a diversity of approaches and analyses. 
Over the past year there have been a number of comparable models released with very permissive licenses, so using a more strict license for our work would not remove the overall risk in the field. 
We believe this trade-off on the side of being more open is the best option.

\section*{Acknowledgments}
 \label{sec:ack}
 \olmo would not have been possible without the support of many individuals and institutions.
 The experimental components of this work were made possible through a partnership with AMD and CSC, enabling use of the LUMI supercomputer, and Kempner Institute at Harvard University.
 We thank Jonathan Frankle and the team at MosaicML (now Databricks) for sharing their experiences with FSDP, and building the code base that OLMo is based on.
 We thank our teammates Taira Anderson, Michelle Benedict, Jon Borchardt, Evie Cheng, Arnavi Chheda, Johann Dahm, Matt Latzke, Kelsey MacMillan, Aaron Sarnat, Carissa Schoenick, Sam Skjonsberg, Michael Schmitz, Michael Wilson, Caitlin Wittlif, and the entire IT team, for their help with the website, design, internal and external communications, budgeting, and other activities that supported smooth progress on this project.
 Finally, we also express gratitude for the helpful discussions and feedback from our teammates at AI2 and close collaborators, including Prithviraj (Raj) Ammanabrolu, Peter Clark, Nicole DeCario, Doug Downey, Ali Farhadi, Ian Ferreira, Väinö Hatanpää, Sham M. Kakade, Julien Launay, Sydney Levine, Pekka Manninen, Franzi Roessner, Maarten Sap, Ludwig Schmidt, Yulia Tsvetkov, and Daniel S. Weld.

\bibliography{references,custom}
\bibliographystyle{acl_natbib}

\newpage
\appendix

\input{appendix}

\end{document}

%% file: appendix.tex
\section{Training Settings}
\label{sec:training_details}
Table~\ref{tab:llm_arch_adamw} summarizes the model architecture and the optimizer parameters of \olmo-7B as well as recent similar-sized models.

\begin{table*}[t]
    \centering
    \small
    \begin{tabular}{l|lllll}
        \toprule
        & \textbf{\olmo-7B} & \textbf{LLaMA2-7B} & \textbf{OpenLM-7B} & \textbf{Falcon-7B} & \textbf{PaLM-8B} \\
        \hline
       Dimension & 4096 & 4096 & 4096 & 4544 & 4096 \\
        Num heads & 32 & 32 & 32 & 71 & 16 \\
        Num layers & 32 & 32 & 32 & 32 & 32 \\
        MLP ratio & $\sim$8/3 & $\sim$8/3 & $\sim$8/3 & 4 & 4 \\
        Layer norm type & non-parametric & RMSNorm & parametric & parametric & parametric \\
        Positional embeddings & RoPE & RoPE & RoPE & RoPE & RoPE \\
        Attention variant & full & GQA & full & MQA & MQA \\
        Biases & none & none & in LN only & in LN only & none \\
        Block type & sequential & sequential & sequential & parallel & parallel \\
        Activation & SwiGLU & SwiGLU & SwiGLU & GeLU & SwiGLU \\
        Sequence length & 2048 & 4096 & 2048 & 2048 & 2048 \\
        Batch size (instances) & 2160 & 1024 & 2048 & 2304 & 512 \\
        Batch size (tokens) & $\sim$4M & $\sim$4M & $\sim$4M & $\sim$4M & $\sim$1M \\
        Weight tying & no & no & no & no & yes \\
        \hline
        Warmup steps & 5000 & 2000 & 2000 & 1000 \\
        Peak LR & \texttt{3.0E-04} & \texttt{3.0E-04} & \texttt{3.0E-04} & \texttt{6.0E-04} \\
        Minimum LR & \texttt{3.0E-05} & \texttt{3.0E-05} & \texttt{3.0E-05} & \texttt{1.2E-05} \\
        Weight decay & 0.1 & 0.1 & 0.1 & 0.1 \\
        Beta1 & 0.9 & 0.9 & 0.9 & 0.99 \\
        Beta2 & 0.95 & 0.95 & 0.95 & 0.999 \\
        Epsilon & \texttt{1.0E-05} & \texttt{1.0E-05} & \texttt{1.0E-05} & \texttt{1.0E-05} \\
        LR schedule & linear & cosine & cosine & cosine \\
        Gradient clipping & global 1.0 & global 1.0 & global 1.0 & global 1.0 \\
        Gradient reduce dtype & FP32 & FP32 & FP32 & BF16 \\
        Optimizer state dtype & FP32 & most likely FP32 & FP32 & FP32 \\
        \bottomrule
    \end{tabular}
    \caption{LM architecture and optimizer comparison at the 7--8B scale. In the ``layer norm type" row, ``parametric" and ``non-parametric" refer to the usual layer norm implementation with and without adaptive gain and bias, respectively. All models are trained using AdamW.}
    \label{tab:llm_arch_adamw}
\end{table*}

\section{Power Consumption and Carbon Footprint}
\label{sec:carbon}

Following previous literature \citep{strubell-etal-2019-energy, patterson2021carbon, wu2022sustainable, dodge2022measuring}, we estimate the total energy consumed and carbon released while pretraining our models by calculating the total power consumption required for training, and then multiplying it by the carbon emission intensity of the power grid where the model was trained. While reporting these operational emissions is standard practice, it does not account for other sources of emissions such as the embodied emissions due to the manufacturing, transportation, and disposal of hardware and datacenter infrastructure, lifetime operational emissions due to use, rebound effects, or other environmental impacts such as water consumption or mining. Thus our estimates should be viewed as lower bounds.

We calculate the total power consumption for our models by measuring the power consumption of a single node every 25ms, calculating an average across the entire training run, and multiplying by the total number of nodes. We then account for the energy efficiency of the data center by multiplying the previous total by a power usage effectiveness (PUE) factor, which we set to 1.1, representing a conservative 10\% energy consumption overhead typical of energy efficient datacenters.\footnote{\protect\url{https://www.nrel.gov/computational-science/measuring-efficiency-pue.html}}\footnote{\protect\url{https://www.google.com/about/datacenters/efficiency/}} We estimate that pretraining our 7B models consumed \textbf{239 MWh} of energy.

To calculate carbon emissions, we multiply the total power consumption by a carbon intensity factor, measured in kg CO$_2$ emitted per KWh, based on the physical location of the data center where each model was trained. The model trained on A100-40GB GPUs was trained in Australia, so we assume a carbon intensity factor of 0.610, the national average for Australia in 2022.\footnote{\protect\url{https://www.cleanenergyregulator.gov.au/Infohub/Markets/Pages/qcmr/december-quarter-2022/Emissions-Reduction.aspx}} The model trained on MI250X GPUs was trained in the LUMI supercomputer, which runs on 100\% renewable, carbon-neutral energy, so we assume a carbon intensity factor of 0. LUMI is powered entirely by hydroelectric power and some sources \citep{Ubierna2022} measure the carbon intensity factor of hydroelectric power to be 0.024, which would imply total carbon emissions of 3.54 tCO$_2$eq.{\footnote{\url{https://www.lumi-supercomputer.eu}}} However, we rely on the official LUMI data for our calculations, and thus we estimate total pretraining emissions of \textbf{69.78 tCO$_2$eq}.\footnote{These metrics were in part collected using Carbonara's AI agent and monitoring platform. Learn more at: \url{https://trycarbonara.com}} In Table \ref{tab:co2-emissions} we compare our models with other previously released models based on publicly available information.

We hope that openly releasing our models can reduce future emissions by allowing others to avoid the need to pretrain models from scratch, and give insights into the true cost of developing state of the art models. We also highlight that our estimates are lower bounds, because they do not include other critical pieces of development such as debugging, hyperparameter tuning, and downtime.

\begin{table*}
\begin{tabular}{l|cccccc}
\toprule
 & GPU Type & \begin{tabular}[c]{@{}c@{}}GPU Power\\ Consumption\\(MWh)\end{tabular} & \begin{tabular}[c]{@{}c@{}}Power\\ Usage\\ Effectiveness\end{tabular} & \begin{tabular}[c]{@{}c@{}}Carbon\\ Intensity\\ (kg CO$_2$e/KWh)\end{tabular} & \begin{tabular}[c]{@{}c@{}}Carbon\\ Emissions\\ (tCO$_2$eq)\end{tabular} \\ \hline
\textbf{Gopher-280B} & TPU v3 & 1,066 & 1.08 & 0.330 & 380 \\
\textbf{BLOOM-176B} & A100-80GB & 433 & 1.2 & 0.057 & 30 \\
\textbf{OPT-175B} & A100-80GB & 324 & 1.1 & 0.231 & 82 \\
\textbf{T5-11B} & TPU v3 & 77 & 1.12 & 0.545 & 47 \\
\textbf{LLaMA-7B} & A100-80GB & 33 & 1.1 & 0.385 & 14 \\
\textbf{LLaMA2-7B} & A100-80GB & 74 & 1.1 & 0.385 & 31 \\
\rowcolor{olmocolor}
[\dimexpr\tabcolsep+0.1pt\relax] 
\textbf{OLMo-7B} & MI250X & 135 & 1.1 & 0.000* & 0* \\
\hline  
\rowcolor{olmocolor}
[\dimexpr\tabcolsep+0.1pt\relax]
\textbf{OLMo-7B} & A100-40GB & 104 & 1.1 & 0.610 & 70 \\
\arrayrulecolor{black}\bottomrule
\end{tabular}
\caption{CO$_2$ emissions during pretraining. We estimate the total carbon emissions for various models using publicly available data on PUE, carbon intensity of local power grid, and reported power consumption. Numbers for Gopher-280B \citep{rae2022scaling}, BLOOM-176B \citep{luccioni2022estimating}, OPT-175B \citep{zhang2022opt}, T5-11B \citep{patterson2021carbon}, LLaMA \citep{touvron2023llama1}, and LLaMA2 \citep{touvron2023llama2} are taken from their respective papers. See Section \ref{sec:carbon} for details on how tCO2eq was calculated.\\
\addtocounter{footnote}{-1}* LUMI runs entirely on hydroelectric power\protect\footnotemark and some estimates \citep{Ubierna2022} measure the intensity factor of hydroelectric power to be 0.024, implying total emissions of 3.54 tCO$_2$eq.}
\label{tab:co2-emissions}
\end{table*}

\section{Additional Evaluation}
\label{sec:add_eval}

\begin{figure*}[t]
    \centering
    \includegraphics[width=\textwidth]{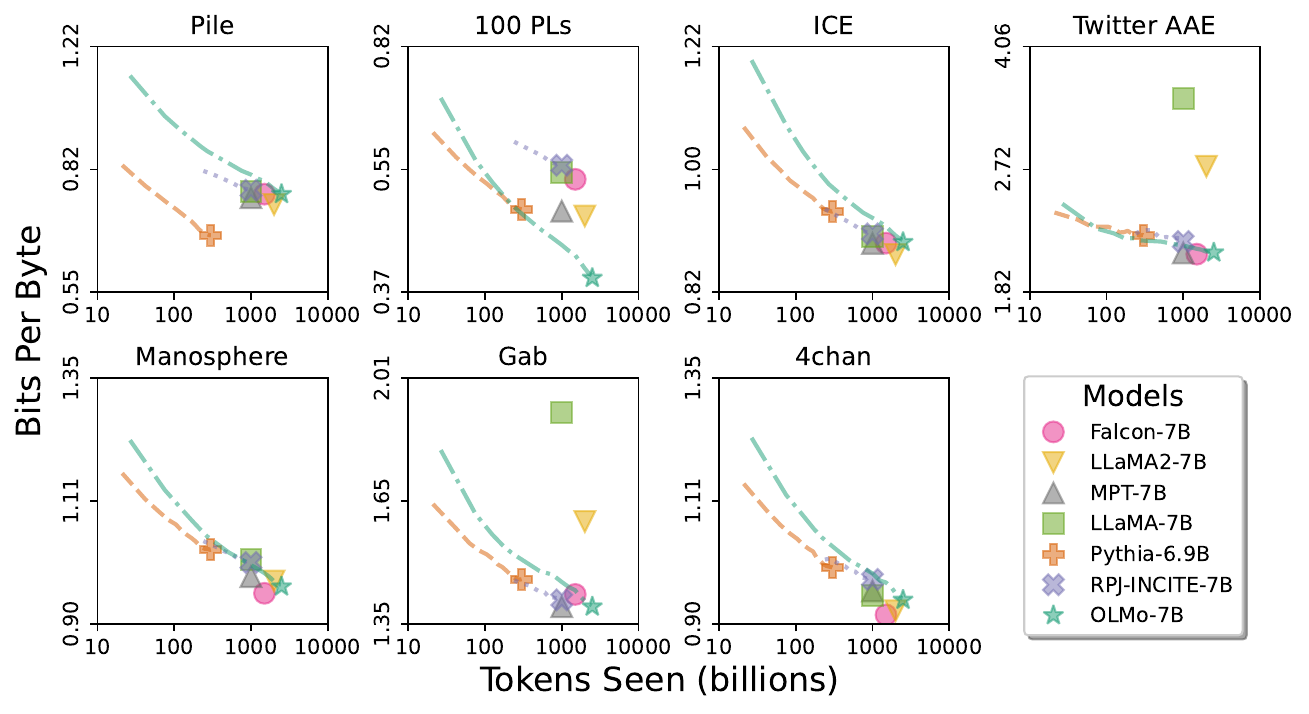}
    \caption{Bits per byte for each of the 7 remaining Paloma data sources not aggregated in Figure~\ref{fig:paloma_bpb_over_tokens_seen_per_source_clean}.
    }
    \label{fig:paloma_bpb_over_tokens_seen_per_source_not_clean}
\end{figure*}

\paragraph{Additional perplexity results} In Figure~\ref{fig:paloma_bpb_over_tokens_seen_per_source_not_clean} we provide results for each of the 7 data sources in Paloma \citep{magnusson2023paloma} that are excluded from the combined metric in Figure~\ref{fig:paloma_bpb_over_tokens_seen_per_source_clean}. Some of these sources such as Pile \citep{gao2020pile} and ICE \citep{GREENBAUM_1996} are not publicly available at this time. Dolma 100 Programming Languages \citep{dolma} consists of code data that is not supported by the decontamination approach used in Paloma. TwitterAAE \citep{blodgett-etal-2016-demographic}, along with ICE, are datasets for targeted analyses of disparities in performance between different dialects and as such should be evaluated separately. And finally, the Manosphere, Gab, and 4chan corpora \citep{Horta_Ribeiro_2021, zannettougab, papasavva_2020} are intended to examine model fit to language from fringe online communities that are studied for prevalent hate speech and toxicity. Thus minimizing perplexity on these fringe corpora is not always desirable.

One notable result here is that \model is much farther ahead of the other models on Dolma 100 Programming Languages (100 PLs). Note that this effect may be due in part to underestimation from contamination, as decontaminating code data is beyond the scope of the method in Paloma. At the same time other models that are trained on code data from GitHub such as RPJ-INCITE-7B, that are just as likely to have contamination, fair much worse. Another factor then is that \model trains on code data with exactly the same post-processing as that in 100 PLs while the code data in other models will have been processed differently. Similarly, Pile evaluation demonstrates these in-distribution and potential contamination effects as Pythia-6.9B achieves top performance despite being trained on almost an order of magnitude fewer tokens than \model.

The results on the remaining 5 targeted sources should be interpreted with care, as Paloma often finds that perplexity on these sources is dominated by superficial features such as low average document length rather than fit to that which would actually be salient to members of these speech communities. TwitterAAE and Gab have among the shortest documents in Paloma contributing to unusually high bits per byte in this figure. Other than these two, the models are notably very closely grouped in a data scaling trend in ICE, Manosphere, and 4chan.

\paragraph{Additional end-task results} Next, in Table~\ref{table:additional-end-task}, we provide results from zero-shot evaluation of \model on 6 additional end-tasks apart from the 8 in our core evaluation suite. These tasks are \texttt{headqa\_en}~\citep{head-qa}, \texttt{logiqa}~\citep{logi-qa}, \texttt{mrpc}~\citep{mrpc}, \texttt{qnli}~\citep{glue}, \texttt{wic}~\citep{wic}, and \texttt{wnli}~\citep{glue}.

\begin{table*}[!t]
\centering 
\begin{tabular}{@{}c|cccccc|c
}
\toprule
& headqa\_en & logiqa & mrpc & qnli & wic & wnli & avg. \\
\hline
\textbf{Falcon-7B}     & 38.6 & 23.7 & 62.8 & 49.8 & 49.5 & 47.9 & 45.4 \\
\textbf{LLaMA-7B}      & 38.7 & 19.5 & 68.6 & 50.1 & 49.1 & 52.1 & 46.4 \\
\textbf{LLaMA2-7B}     & 39.5 & 26.1 & 69.1 & 49.4 & 49.8 & 45.1 & 46.5 \\
\textbf{MPT-7B}        & 37.4 & 22.9 & 67.7 & 52.1 & 48.1 & 47.9 & 46.0 \\
\textbf{Pythia-6.9B}   & 40.1 & 21.5 & 65.4 & 53.8 & 55.0 & 38.0 & 45.6 \\
\textbf{RPJ-INCITE-7B} & 36.9 & 27.8 & 58.8 & 53.8 & 48.9 & 57.8 & 47.3 \\
\rowcolor{olmocolor}
[\dimexpr\tabcolsep+0.1pt\relax]
\textbf{OLMo-7B}       & 37.3 & 23.4 & 68.4 & 49.1 & 50.2 & 56.3 & 47.5 \\
\bottomrule
\end{tabular}
\caption{Zero-shot evaluation of \model on 6 additional end-tasks apart from the 8 present in our core evaluation suite. Once again, we compare \model to 6 other model checkpoints which are publicly available. We find that \model outperforms the other models on aggregate taken over 6 additional end-tasks from this table, however these tasks were also found to provide limited signal during training (see Figure~\ref{fig:additional-task-acc-progression}).}
\label{table:additional-end-task}
\end{table*}

\begin{figure*}[!t]
\centering
\makebox[\columnwidth][c]{
\includegraphics[width=\textwidth]{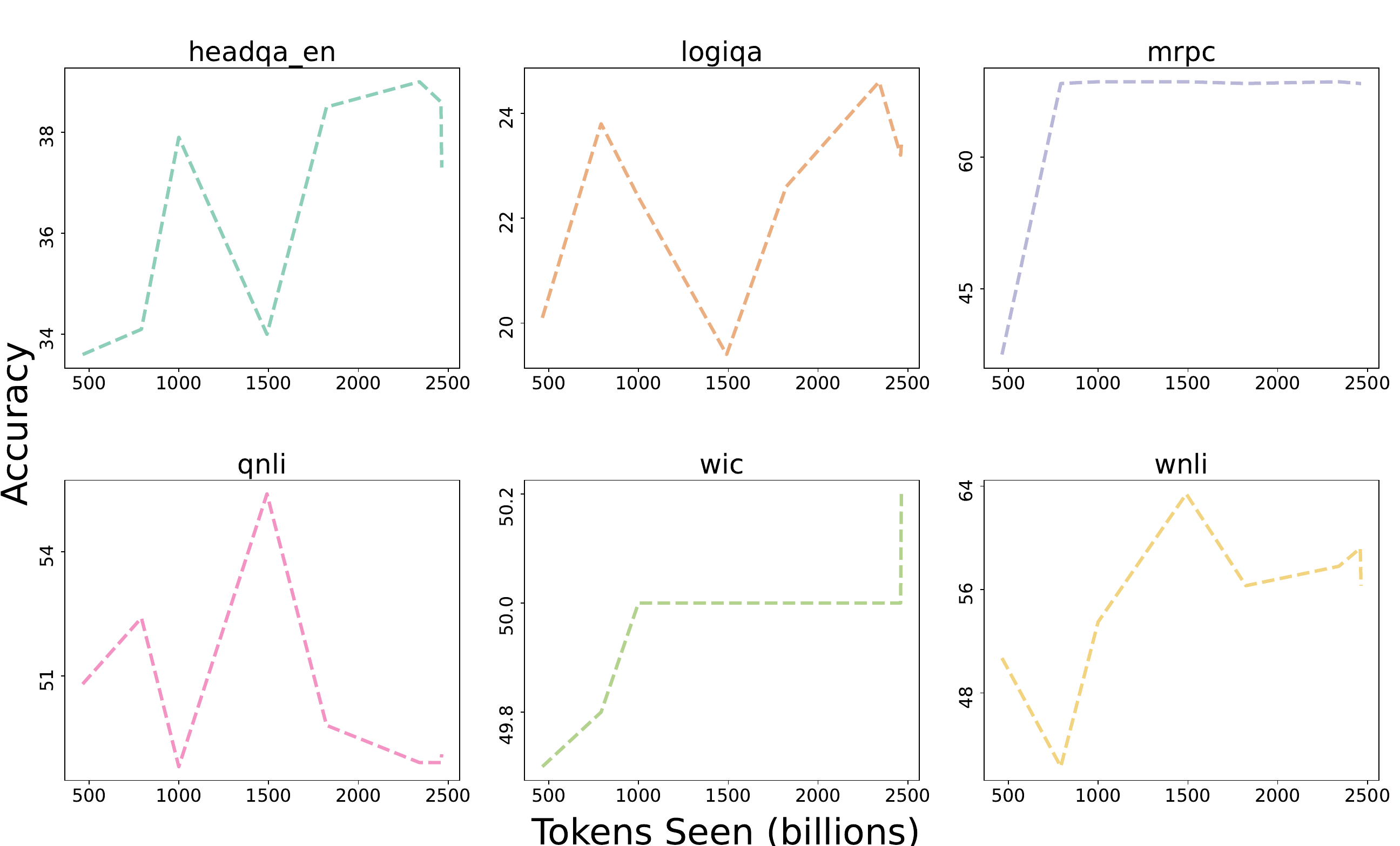}
}
\caption{Accuracy score progression of \model on 6 additional end-tasks. The performance of these additional end-tasks was unstable and provided limited signal during model development.}
\label{fig:additional-task-acc-progression}
\end{figure*}

We note, however, that in contrast to our core evaluation set described in Section~\ref{sec:downstream_evaluation}, we found these additional end-tasks to have less stable performance during model development, and to provide a limited signal.  This is illustrated in Figure~\ref{fig:additional-task-acc-progression}, where we see the progress of task performance throughout training to be more random (compare with the more stable upward trends in Figure~\ref{fig:core-tasks-progression}). While tasks such as \texttt{mrpc} and \texttt{wic} appear more stable, they offered additional difficulties related to performance being tied to random chance (e.g., \texttt{wic}) or the tendency of models to make spurious predictions (e.g., always predicting a single label) that either inflate or deflate performance due to dataset class imbalances (e.g., \texttt{mrpc}). We therefore caution against relying too heavily on these tasks when measuring model performance throughout training and comparing models.

\section{Adaptation Training Details}
\label{app:adaptation_training_details}
We use the following hyperparameters when instruction tuning \olmo. These were chosen through small pilot experiments.

\begin{itemize}
\setlength{\itemsep}{0pt}
    \item Learning rate: $2 \times 10^{-6}$
    \item Epochs: 3
    \item Warmup: Linear warmup for the first 3\% of total training time, and then linear cooldown to a learning rate of 0 over the remaining steps.
    \item Weight decay: 0
    \item Gradient clipping: 0
    \item Maximum sequence length: 2048
    \item Data: \tulu V2 SFT mix, resplit such that long conversations are split into 2048-token chunks and replacing the hardcoded split with data about \olmo. Data is publically available.\footnote{\url{https://huggingface.co/datasets/allenai/tulu-v2-sft-mixture-olmo-2048}}
\end{itemize}

After instruction finetuning, we then use the following hyperparameters for DPO training, following \citet{ivison2023camels}:

\begin{itemize}
\setlength{\itemsep}{0pt}
    \item Learning rate: $5 \times 10^{-7}$
    \item $\beta$: 0.1
    \item Epochs: 3
    \item Warmup: Linear warmup for the first 10\% of total training time, and then linear cooldown to a learning rate of 0 over the remaining steps.
    \item Weight decay: 0
    \item Gradient clipping: 0
    \item Maximum sequence length: 2048
    \item Data: A modified form of UltraFeedback~\citep{cui2023ultrafeedback}, with TruthfulQA prompts removed. We used the `fixed' variant released by Argilla, which uses the average of GPT-generated aspect-based scores to determine chosen and rejected pairs.\footnote{\url{https://huggingface.co/datasets/argilla/ultrafeedback-binarized-preferences-cleaned}}
\end{itemize}

\section{Adaptation Evaluation and Model details}
\label{app:adaptation-evals-description}
We choose the models in Table~\ref{table:results-adaptation} by choosing the `canonical' best versions (that is, the best instruction-tuned or otherwise adapted models released by the same organisation) of the base models we compare against in Table~\ref{table:results-end-task}. We additionally compare to \tulu 2 to show the current best models trained using the \tulu mix used to finetune \olmo. We display evaluations on MMLU, AlpacaEval, ToxiGen, and Truthfulness to focus on displaying how instruction tuning can generally help capabilities (MMLU), how the models perform in an open-ended chat setting (AlpacaEval), and to test how instruction tuning aids in model safety and truthfulness (AlpacaEval, ToxiGen). We additionally report \olmo's performance over the entire \tulu evaluation suite in Table~\ref{table:results-adaptation-full}.

\begin{table*}[t]
\adjustbox{max width=2\columnwidth}{
\begin{tabular}{@{}l|cccccccc
}
\toprule
\textbf{Model} & \textbf{MMLU}   & \textbf{GSM8k}      & \textbf{BBH}        & \textbf{TydiQA} & \textbf{Codex-Eval} & \textbf{AlpacaEval} & \textbf{ToxiGen} & \textbf{TruthfulQA} \\
& \textbf{0-shot} & \textbf{8-shot CoT} & \textbf{3-shot CoT} & \textbf{1-shot} & \textbf{Pass@10}    & \textbf{\%win}       & \textbf{\% Toxic} & \textbf{\% Info + True}  \\
\hline
\textbf{\model}        & 28.3 & 8.5 & 31.7 & 32.3 & 21.4 & - & 81.4 & 31.6 \\
\textbf{+SFT}        & 47.3 & 15.5 & 36.9 & 35.2 & 28.6 & 57.0 & 14.4 & 41.2 \\
\textbf{+SFT+DPO}        & 46.1 & 11.0 & 35.8 & 21.7 & 27.8 & 69.3  & 1.7 & 52.0 \\
\bottomrule
\end{tabular}
}
\vspace{1em}
\caption{Evaluation of \model models before and after instruction finetuning and DPO training on the full \tulu evaluation suite. Lower is better for ToxiGen and higher is better for other metrics.}
\label{table:results-adaptation-full}
\end{table*}

We provide a brief description of each model evaluated in Table~\ref{table:results-adaptation} below. For all models, we use the provided chat template for prompt formatting when available. 
\begin{itemize}[leftmargin=*]
    \item MPT Chat: A version of MPT 7B finetuned on the ShareGPT-Vicuna~\citep{vicuna2023}, HC3~\citep{guo-etal-2023-hc3}, Alpaca~\citep{alpaca}, HH-RLHF~\citep{bai2022training}, and Evol-Instruct~\citep{xu2024wizardlm} datasets. Retrieved from \url{https://huggingface.co/mosaicml/mpt-7b-chat}.
    \item Falcon Instruct: A version of Falcon 7B finetuned on the Baize~\citep{xu2023baize}, GPT4All~\citep{gpt4all}, GPTeacher~\citep{gpteacher}, and Refined-Web English~\citep{penedo2023therd} datasets. Retrieved from \url{https://huggingface.co/tiiuae/falcon-7b-instruct}.
    \item RPJ-INCITE Chat: A version of RPJ-INCITE 7B finetuned on the OASST1~\citep{kopf2023openassistant} and Dolly V2~\citep{DatabricksBlog2023DollyV2} datasets. Retrieved from \url{https://huggingface.co/togethercomputer/RedPajama-INCITE-7B-Chat}.
    \item Llama-2 Chat: A version of Llama 2 7B finetuned on a mixture of instruction datasets and further trained with RLHF. We refer the reader to \citet{touvron2023llama2} for further details.
    \item \tulu 2: A version of Llama 2 7B finetuned on a mixture of instruction datasets (the \tulu 2 mix). We refer the reader to \citet{ivison2023camels} for further details.
    \item \tulu 2+DPO: \tulu 2 further trained with DPO on the UltraFeedback dataset~\citep{cui2023ultrafeedback}. We refer the reader to \citet{ivison2023camels} for further details.
    \item OLMo+SFT: A version of \olmo 7B fintuned on the same data as \tulu 2.
    \item OLMo+SFT+DPO: OLMo+SFT further trained with DPO on the UltraFeedback dataset~\citep{cui2023ultrafeedback}.
\end{itemize}

We additionally provide a brief description of each evaluation setting from Table~\ref{table:results-adaptation}:
\begin{itemize}[leftmargin=*]
    \item \textbf{MMLU}: We use the official MMLU~\citep{hendryckstest2021} evaluation script and prompts available at \url{https://github.com/hendrycks/test}, with modifications to allow for batch processing. We evaluate using 0 few-shot examples, following the original setup of MMLU. We report average accuracy across test examples.
    \item \textbf{ToxiGen}: We follow the setup in \citet{touvron2023llama2}, but use the original set of prompts from \citet{hartvigsen2022toxigen}, which are designed to elicit toxic generations for certain groups. We take only the prompts designed to produce toxic language (`hateful' prompts) and use 500 prompts per group to reduce evaluation costs. For base language models, we pass in the original ToxiGen prompts unchanged and greedily decode up to the first new line (or a maximum of 512 tokens). For instruction-tuned models, we place the prompt in the corresponding template, and ask the model to complete the prompt, until the model generates a stop token (or a maximum of 512 tokens). We pass the generated text into a roberta-large model trained to detect toxic content finetuned as part of \citet{hartvigsen2022toxigen}.\footnote{\url{https://huggingface.co/tomh/toxigen\_roberta}} We then report the percentage of generations deemed toxic by the classifier.
    \item \textbf{TruthfulQA}: Following \citet{touvron2023llama2}, we mainly use the generation setting of TruthfulQA \citep{lin2022truthfulqa}. The TruthfulQA dataset contains 818 questions, which are used to prompt the tested model to generate answers. We use the default QA prompt format with 6 in-context QA examples. We follow the official script in their official implemention\footnote{\url{https://github.com/sylinrl/TruthfulQA/}} to do greedy decoding and answer postprocessing. We train two LLaMA 2-based classifiers for judging the truthfulness and informativeness of the model response, due to the deprecation of GPT-3 making exact replication of the original TruthfulQA evaluation infeasible. We find that the LLaMA 2 judges are generally able to match the performance of the original GPT-3-based judges used by \citet{lin2022truthfulqa}. We report the rate of the responses being truthful and informative (\% Informative and Truthful) following \citet{touvron2023llama2}. We only report the \% Informative and Truthful as our primary metric.
    \item \textbf{AlpacaEval}: We use the package provided by \citet{alpaca_eval}, following the default setup which asks the evaluated model to generate responses for 805 prompts and employ GPT-4 to compare the response with \instructgpt{003}. We employ the ``alpaca\_eval\_gpt4'' annotator. We allow the evaluated model to generate up to 2048 tokens, without specifying special stop sequences. The reported win-rate is the percentage of model generations that GPT-4 reports as being preferred over the generations from \instructgpt{003}.
\end{itemize}

%% file: acl2023.bbl
\begin{thebibliography}{91}
\expandafter\ifx\csname natexlab\endcsname\relax\def\natexlab#1{#1}\fi

\bibitem[{Abbas et~al.(2023)Abbas, Tirumala, Simig, Ganguli, and Morcos}]{abbas2023semdedup}
Amro Abbas, Kushal Tirumala, D{\'a}niel Simig, Surya Ganguli, and Ari~S Morcos. 2023.
\newblock \href {https://arxiv.org/abs/2303.09540} {Semdedup: Data-efficient learning at web-scale through semantic deduplication}.
\newblock \emph{arXiv preprint arXiv:2303.09540}.

\bibitem[{Almazrouei et~al.(2023)Almazrouei, Alobeidli, Alshamsi, Cappelli, Cojocaru, Hesslow, Launay, Malartic, Mazzotta, Noune, Pannier, and Penedo}]{Falcon}
Ebtesam Almazrouei, Hamza Alobeidli, Abdulaziz Alshamsi, Alessandro Cappelli, Ruxandra-Aim{\'e}e Cojocaru, Daniel Hesslow, Julien Launay, Quentin Malartic, Daniele Mazzotta, Badreddine Noune, Baptiste Pannier, and Guilherme Penedo. 2023.
\newblock \href {https://api.semanticscholar.org/CorpusID:265466629} {The falcon series of open language models}.
\newblock \emph{ArXiv}, abs/2311.16867.

\bibitem[{Anand et~al.(2023)Anand, Nussbaum, Duderstadt, Schmidt, and Mulyar}]{gpt4all}
Yuvanesh Anand, Zach Nussbaum, Brandon Duderstadt, Benjamin Schmidt, and Andriy Mulyar. 2023.
\newblock Gpt4all: Training an assistant-style chatbot with large scale data distillation from gpt-3.5-turbo.
\newblock \url{https://github.com/nomic-ai/gpt4all}.

\bibitem[{Ba et~al.(2016)Ba, Kiros, and Hinton}]{Ba2016LayerNorm}
Jimmy Ba, Jamie~Ryan Kiros, and Geoffrey~E. Hinton. 2016.
\newblock \href {https://api.semanticscholar.org/CorpusID:8236317} {Layer normalization}.
\newblock \emph{ArXiv}, abs/1607.06450.

\bibitem[{Bai et~al.(2022)Bai, Jones, Ndousse, Askell, Chen, DasSarma, Drain, Fort, Ganguli, Henighan, Joseph, Kadavath, Kernion, Conerly, El-Showk, Elhage, Hatfield-Dodds, Hernandez, Hume, Johnston, Kravec, Lovitt, Nanda, Olsson, Amodei, Brown, Clark, McCandlish, Olah, Mann, and Kaplan}]{bai2022training}
Yuntao Bai, Andy Jones, Kamal Ndousse, Amanda Askell, Anna Chen, Nova DasSarma, Dawn Drain, Stanislav Fort, Deep Ganguli, Tom Henighan, Nicholas Joseph, Saurav Kadavath, Jackson Kernion, Tom Conerly, Sheer El-Showk, Nelson Elhage, Zac Hatfield-Dodds, Danny Hernandez, Tristan Hume, Scott Johnston, Shauna Kravec, Liane Lovitt, Neel Nanda, Catherine Olsson, Dario Amodei, Tom Brown, Jack Clark, Sam McCandlish, Chris Olah, Ben Mann, and Jared Kaplan. 2022.
\newblock \href {http://arxiv.org/abs/2204.05862} {Training a helpful and harmless assistant with reinforcement learning from human feedback}.

\bibitem[{Bengio et~al.(2003)Bengio, Ducharme, Vincent, and Janvin}]{Bengio2003ANP}
Yoshua Bengio, R{\'e}jean Ducharme, Pascal Vincent, and Christian Janvin. 2003.
\newblock \href {https://api.semanticscholar.org/CorpusID:221275765} {A neural probabilistic language model}.
\newblock \emph{J. Mach. Learn. Res.}, 3:1137--1155.

\bibitem[{Biderman et~al.(2023)Biderman, Schoelkopf, Anthony, Bradley, O'Brien, Hallahan, Khan, Purohit, Prashanth, Raff, Skowron, Sutawika, and Van Der~Wal}]{pmlr-v202-biderman23a}
Stella Biderman, Hailey Schoelkopf, Quentin~Gregory Anthony, Herbie Bradley, Kyle O'Brien, Eric Hallahan, Mohammad~Aflah Khan, Shivanshu Purohit, Usvsn~Sai Prashanth, Edward Raff, Aviya Skowron, Lintang Sutawika, and Oskar Van Der~Wal. 2023.
\newblock \href {https://proceedings.mlr.press/v202/biderman23a.html} {Pythia: A suite for analyzing large language models across training and scaling}.
\newblock In \emph{Proceedings of the 40th International Conference on Machine Learning}, volume 202 of \emph{Proceedings of Machine Learning Research}, pages 2397--2430. PMLR.

\bibitem[{BigScience et~al.(2022)BigScience, Scao, Fan, Akiki, Pavlick, Ili{\'c}, Hesslow, Castagn{\'e}, Luccioni, Yvon et~al.}]{workshop2022bloom}
BigScience, Teven~Le Scao, Angela Fan, Christopher Akiki, Ellie Pavlick, Suzana Ili{\'c}, Daniel Hesslow, Roman Castagn{\'e}, Alexandra~Sasha Luccioni, Fran{\c{c}}ois Yvon, et~al. 2022.
\newblock Bloom: A 176b-parameter open-access multilingual language model.
\newblock \emph{arXiv preprint arXiv:2211.05100}.

\bibitem[{Bisk et~al.(2020)Bisk, Zellers, Gao, Choi et~al.}]{bisk2020piqa}
Yonatan Bisk, Rowan Zellers, Jianfeng Gao, Yejin Choi, et~al. 2020.
\newblock \href {https://ojs.aaai.org/index.php/AAAI/article/view/6239} {Piqa: Reasoning about physical commonsense in natural language}.
\newblock In \emph{Proceedings of the AAAI conference on artificial intelligence}, volume~34, pages 7432--7439.

\bibitem[{Black et~al.(2022)Black, Biderman, Hallahan, Anthony, Gao, Golding, He, Leahy, McDonell, Phang, Pieler, Prashanth, Purohit, Reynolds, Tow, Wang, and Weinbach}]{gpt-neox-20b}
Sid Black, Stella Biderman, Eric Hallahan, Quentin Anthony, Leo Gao, Laurence Golding, Horace He, Connor Leahy, Kyle McDonell, Jason Phang, Michael Pieler, USVSN~Sai Prashanth, Shivanshu Purohit, Laria Reynolds, Jonathan Tow, Ben Wang, and Samuel Weinbach. 2022.
\newblock \href {https://arxiv.org/abs/2204.06745} {{GPT-NeoX-20B}: An open-source autoregressive language model}.
\newblock In \emph{Proceedings of the ACL Workshop on Challenges \& Perspectives in Creating Large Language Models}.

\bibitem[{Blodgett et~al.(2016)Blodgett, Green, and O{'}Connor}]{blodgett-etal-2016-demographic}
Su~Lin Blodgett, Lisa Green, and Brendan O{'}Connor. 2016.
\newblock \href {https://doi.org/10.18653/v1/D16-1120} {Demographic dialectal variation in social media: A case study of {A}frican-{A}merican {E}nglish}.
\newblock In \emph{Proceedings of the 2016 Conference on Empirical Methods in Natural Language Processing}, pages 1119--1130, Austin, Texas. Association for Computational Linguistics.

\bibitem[{Brown et~al.(2020)Brown, Mann, Ryder, Subbiah, Kaplan, Dhariwal, Neelakantan, Shyam, Sastry, Askell, Agarwal, Herbert-Voss, Krueger, Henighan, Child, Ramesh, Ziegler, Wu, Winter, Hesse, Chen, Sigler, Litwin, Gray, Chess, Clark, Berner, McCandlish, Radford, Sutskever, and Amodei}]{Brown2020LanguageMA}
Tom~B. Brown, Benjamin Mann, Nick Ryder, Melanie Subbiah, Jared Kaplan, Prafulla Dhariwal, Arvind Neelakantan, Pranav Shyam, Girish Sastry, Amanda Askell, Sandhini Agarwal, Ariel Herbert-Voss, Gretchen Krueger, T.~J. Henighan, Rewon Child, Aditya Ramesh, Daniel~M. Ziegler, Jeff Wu, Clemens Winter, Christopher Hesse, Mark Chen, Eric Sigler, Mateusz Litwin, Scott Gray, Benjamin Chess, Jack Clark, Christopher Berner, Sam McCandlish, Alec Radford, Ilya Sutskever, and Dario Amodei. 2020.
\newblock \href {https://api.semanticscholar.org/CorpusID:218971783} {Language models are few-shot learners}.
\newblock \emph{ArXiv}, abs/2005.14165.

\bibitem[{Chiang et~al.(2023)Chiang, Li, Lin, Sheng, Wu, Zhang, Zheng, Zhuang, Zhuang, Gonzalez, Stoica, and Xing}]{vicuna2023}
Wei-Lin Chiang, Zhuohan Li, Zi~Lin, Ying Sheng, Zhanghao Wu, Hao Zhang, Lianmin Zheng, Siyuan Zhuang, Yonghao Zhuang, Joseph~E. Gonzalez, Ion Stoica, and Eric~P. Xing. 2023.
\newblock \href {https://lmsys.org/blog/2023-03-30-vicuna/} {Vicuna: An open-source chatbot impressing gpt-4 with 90\%* chatgpt quality}.

\bibitem[{Chowdhery et~al.(2022)Chowdhery, Narang, Devlin, Bosma, Mishra, Roberts, Barham, Chung, Sutton, Gehrmann, Schuh, Shi, Tsvyashchenko, Maynez, Rao, Barnes, Tay, Shazeer, Prabhakaran, Reif, Du, Hutchinson, Pope, Bradbury, Austin, Isard, Gur-Ari, Yin, Duke, Levskaya, Ghemawat, Dev, Michalewski, Garcia, Misra, Robinson, Fedus, Zhou, Ippolito, Luan, Lim, Zoph, Spiridonov, Sepassi, Dohan, Agrawal, Omernick, Dai, Pillai, Pellat, Lewkowycz, Moreira, Child, Polozov, Lee, Zhou, Wang, Saeta, Diaz, Firat, Catasta, Wei, Meier-Hellstern, Eck, Dean, Petrov, and Fiedel}]{chowdhery2022palm}
Aakanksha Chowdhery, Sharan Narang, Jacob Devlin, Maarten Bosma, Gaurav Mishra, Adam Roberts, Paul Barham, Hyung~Won Chung, Charles Sutton, Sebastian Gehrmann, Parker Schuh, Kensen Shi, Sasha Tsvyashchenko, Joshua Maynez, Abhishek Rao, Parker Barnes, Yi~Tay, Noam Shazeer, Vinodkumar Prabhakaran, Emily Reif, Nan Du, Ben Hutchinson, Reiner Pope, James Bradbury, Jacob Austin, Michael Isard, Guy Gur-Ari, Pengcheng Yin, Toju Duke, Anselm Levskaya, Sanjay Ghemawat, Sunipa Dev, Henryk Michalewski, Xavier Garcia, Vedant Misra, Kevin Robinson, Liam Fedus, Denny Zhou, Daphne Ippolito, David Luan, Hyeontaek Lim, Barret Zoph, Alexander Spiridonov, Ryan Sepassi, David Dohan, Shivani Agrawal, Mark Omernick, Andrew~M. Dai, Thanumalayan~Sankaranarayana Pillai, Marie Pellat, Aitor Lewkowycz, Erica Moreira, Rewon Child, Oleksandr Polozov, Katherine Lee, Zongwei Zhou, Xuezhi Wang, Brennan Saeta, Mark Diaz, Orhan Firat, Michele Catasta, Jason Wei, Kathy Meier-Hellstern, Douglas Eck, Jeff Dean, Slav Petrov, and Noah Fiedel. 2022.
\newblock \href {http://arxiv.org/abs/2204.02311} {Palm: Scaling language modeling with pathways}.

\bibitem[{Chronopoulou et~al.(2022)Chronopoulou, Peters, and Dodge}]{chronopoulou-etal-2022-efficient}
Alexandra Chronopoulou, Matthew Peters, and Jesse Dodge. 2022.
\newblock \href {https://doi.org/10.18653/v1/2022.naacl-main.96} {Efficient hierarchical domain adaptation for pretrained language models}.
\newblock In \emph{Proceedings of the 2022 Conference of the North American Chapter of the Association for Computational Linguistics: Human Language Technologies}, pages 1336--1351, Seattle, United States. Association for Computational Linguistics.

\bibitem[{Chung et~al.(2023)Chung, Constant, Garc{\'i}a, Roberts, Tay, Narang, and Firat}]{chung2023unimaxfa}
Hyung~Won Chung, Noah Constant, Xavier Garc{\'i}a, Adam Roberts, Yi~Tay, Sharan Narang, and Orhan Firat. 2023.
\newblock \href {https://api.semanticscholar.org/CorpusID:258187051} {Unimax: Fairer and more effective language sampling for large-scale multilingual pretraining}.
\newblock \emph{ArXiv}, abs/2304.09151.

\bibitem[{Clark et~al.(2019)Clark, Lee, Chang, Kwiatkowski, Collins, and Toutanova}]{clark2019boolq}
Christopher Clark, Kenton Lee, Ming-Wei Chang, Tom Kwiatkowski, Michael Collins, and Kristina Toutanova. 2019.
\newblock Boolq: Exploring the surprising difficulty of natural yes/no questions.
\newblock \emph{arXiv preprint arXiv:1905.10044}.

\bibitem[{Clark et~al.(2018)Clark, Cowhey, Etzioni, Khot, Sabharwal, Schoenick, and Tafjord}]{clark2018think}
Peter Clark, Isaac Cowhey, Oren Etzioni, Tushar Khot, Ashish Sabharwal, Carissa Schoenick, and Oyvind Tafjord. 2018.
\newblock \href {https://arxiv.org/abs/1803.05457} {Think you have solved question answering? try arc, the ai2 reasoning challenge}.
\newblock \emph{arXiv preprint arXiv:1803.05457}.

\bibitem[{Conover et~al.(2023)Conover, Hayes, Mathur, Xie, Wan, Shah, Ghodsi, Wendell, Zaharia, and Xin}]{DatabricksBlog2023DollyV2}
Mike Conover, Matt Hayes, Ankit Mathur, Jianwei Xie, Jun Wan, Sam Shah, Ali Ghodsi, Patrick Wendell, Matei Zaharia, and Reynold Xin. 2023.
\newblock \href {https://www.databricks.com/blog/2023/04/12/dolly-first-open-commercially-viable-instruction-tuned-llm} {Free dolly: Introducing the world's first truly open instruction-tuned llm}.

\bibitem[{Cui et~al.(2023)Cui, Yuan, Ding, Yao, Zhu, Ni, Xie, Liu, and Sun}]{cui2023ultrafeedback}
Ganqu Cui, Lifan Yuan, Ning Ding, Guanming Yao, Wei Zhu, Yuan Ni, Guotong Xie, Zhiyuan Liu, and Maosong Sun. 2023.
\newblock \href {http://arxiv.org/abs/2310.01377} {Ultrafeedback: Boosting language models with high-quality feedback}.

\bibitem[{Dodge et~al.(2022)Dodge, Prewitt, Combes, Odmark, Schwartz, Strubell, Luccioni, Smith, DeCario, and Buchanan}]{dodge2022measuring}
Jesse Dodge, Taylor Prewitt, Remi Tachet~Des Combes, Erika Odmark, Roy Schwartz, Emma Strubell, Alexandra~Sasha Luccioni, Noah~A. Smith, Nicole DeCario, and Will Buchanan. 2022.
\newblock \href {http://arxiv.org/abs/2206.05229} {Measuring the carbon intensity of ai in cloud instances}.

\bibitem[{Dolan and Brockett(2005)}]{mrpc}
William~B. Dolan and Chris Brockett. 2005.
\newblock \href {https://www.microsoft.com/en-us/research/publication/automatically-constructing-a-corpus-of-sentential-paraphrases/} {Automatically constructing a corpus of sentential paraphrases}.
\newblock In \emph{International Joint Conference on Natural Language Processing}.

\bibitem[{Elazar et~al.(2024)Elazar, Bhagia, Magnusson, Ravichander, Schwenk, Suhr, Walsh, Groeneveld, Soldaini, Singh, Hajishirzi, Smith, and Dodge}]{wimbd}
Yanai Elazar, Akshita Bhagia, Ian~Helgi Magnusson, Abhilasha Ravichander, Dustin Schwenk, Alane Suhr, Evan~Pete Walsh, Dirk Groeneveld, Luca Soldaini, Sameer Singh, Hanna Hajishirzi, Noah~A. Smith, and Jesse Dodge. 2024.
\newblock \href {https://openreview.net/forum?id=RvfPnOkPV4} {What's in my big data?}
\newblock In \emph{The Twelfth International Conference on Learning Representations}.

\bibitem[{Gao et~al.(2020)Gao, Biderman, Black, Golding, Hoppe, Foster, Phang, He, Thite, Nabeshima et~al.}]{gao2020pile}
Leo Gao, Stella Biderman, Sid Black, Laurence Golding, Travis Hoppe, Charles Foster, Jason Phang, Horace He, Anish Thite, Noa Nabeshima, et~al. 2020.
\newblock \href {https://arxiv.org/abs/2101.00027} {The pile: An 800gb dataset of diverse text for language modeling}.
\newblock \emph{arXiv preprint arXiv:2101.00027}.

\bibitem[{Gao et~al.(2023)Gao, Tow, Abbasi, Biderman, Black, DiPofi, Foster, Golding, Hsu, Le~Noac'h, Li, McDonell, Muennighoff, Ociepa, Phang, Reynolds, Schoelkopf, Skowron, Sutawika, Tang, Thite, Wang, Wang, and Zou}]{eval-harness}
Leo Gao, Jonathan Tow, Baber Abbasi, Stella Biderman, Sid Black, Anthony DiPofi, Charles Foster, Laurence Golding, Jeffrey Hsu, Alain Le~Noac'h, Haonan Li, Kyle McDonell, Niklas Muennighoff, Chris Ociepa, Jason Phang, Laria Reynolds, Hailey Schoelkopf, Aviya Skowron, Lintang Sutawika, Eric Tang, Anish Thite, Ben Wang, Kevin Wang, and Andy Zou. 2023.
\newblock \href {https://doi.org/10.5281/zenodo.10256836} {A framework for few-shot language model evaluation}.

\bibitem[{Greenbaum and Nelson(1996)}]{GREENBAUM_1996}
Sidney Greenbaum and Gerald Nelson. 1996.
\newblock \href {https://doi.org/10.1111/j.1467-971x.1996.tb00088.x} {The international corpus of english ({ICE}) project}.
\newblock \emph{World Englishes}, 15(1):3--15.

\bibitem[{Groeneveld et~al.(2023)Groeneveld, Awadalla, Beltagy, Bhagia, Magnusson, Peng, Tafjord, Walsh, Richardson, and Dodge}]{groeneveld2023catwalk}
Dirk Groeneveld, Anas Awadalla, Iz~Beltagy, Akshita Bhagia, Ian Magnusson, Hao Peng, Oyvind Tafjord, Pete Walsh, Kyle Richardson, and Jesse Dodge. 2023.
\newblock \href {https://arxiv.org/abs/2312.10253} {Catwalk: A unified language model evaluation framework for many datasets}.
\newblock \emph{arXiv preprint arXiv:2312.10253}.

\bibitem[{Guo et~al.(2023)Guo, Zhang, Wang, Jiang, Nie, Ding, Yue, and Wu}]{guo-etal-2023-hc3}
Biyang Guo, Xin Zhang, Ziyuan Wang, Minqi Jiang, Jinran Nie, Yuxuan Ding, Jianwei Yue, and Yupeng Wu. 2023.
\newblock How close is chatgpt to human experts? comparison corpus, evaluation, and detection.
\newblock \emph{arXiv preprint arxiv:2301.07597}.

\bibitem[{Gururangan et~al.(2023)Gururangan, Wortsman, Gadre, Dave, Kilian, Shi, Mercat, Smyrnis, Ilharco, Jordan, Heckel, Dimakis, Farhadi, Shankar, and Schmidt}]{open_lm}
Suchin Gururangan, Mitchell Wortsman, Samir~Yitzhak Gadre, Achal Dave, Maciej Kilian, Weijia Shi, Jean Mercat, Georgios Smyrnis, Gabriel Ilharco, Matt Jordan, Reinhard Heckel, Alex Dimakis, Ali Farhadi, Vaishaal Shankar, and Ludwig Schmidt. 2023.
\newblock \href {https://github.com/mlfoundations/open_lm/} {{OpenLM}: a minimal but performative language modeling (lm) repository}.
\newblock GitHub repository.

\bibitem[{Hartvigsen et~al.(2022)Hartvigsen, Gabriel, Palangi, Sap, Ray, and Kamar}]{hartvigsen2022toxigen}
Thomas Hartvigsen, Saadia Gabriel, Hamid Palangi, Maarten Sap, Dipankar Ray, and Ece Kamar. 2022.
\newblock \href {https://arxiv.org/abs/2203.09509} {{TOXIGEN: Controlling Language Models to Generate Implied and Adversarial Toxicity}}.
\newblock In \emph{ACL}.

\bibitem[{Hendrycks et~al.(2021)Hendrycks, Burns, Basart, Zou, Mazeika, Song, and Steinhardt}]{hendryckstest2021}
Dan Hendrycks, Collin Burns, Steven Basart, Andy Zou, Mantas Mazeika, Dawn Song, and Jacob Steinhardt. 2021.
\newblock Measuring massive multitask language understanding.
\newblock \emph{Proceedings of the International Conference on Learning Representations (ICLR)}.

\bibitem[{Ivison et~al.(2023)Ivison, Wang, Pyatkin, Lambert, Peters, Dasigi, Jang, Wadden, Smith, Beltagy, and Hajishirzi}]{ivison2023camels}
Hamish Ivison, Yizhong Wang, Valentina Pyatkin, Nathan Lambert, Matthew Peters, Pradeep Dasigi, Joel Jang, David Wadden, Noah~A. Smith, Iz~Beltagy, and Hannaneh Hajishirzi. 2023.
\newblock \href {http://arxiv.org/abs/2311.10702} {Camels in a changing climate: Enhancing lm adaptation with tulu 2}.

\bibitem[{Jiang et~al.(2024)Jiang, Sablayrolles, Roux, Mensch, Savary, Bamford, Chaplot, Casas, Hanna, Bressand et~al.}]{jiang2024mixtral}
Albert~Q Jiang, Alexandre Sablayrolles, Antoine Roux, Arthur Mensch, Blanche Savary, Chris Bamford, Devendra~Singh Chaplot, Diego de~las Casas, Emma~Bou Hanna, Florian Bressand, et~al. 2024.
\newblock \href {https://arxiv.org/abs/2401.04088} {Mixtral of experts}.
\newblock \emph{arXiv preprint arXiv:2401.04088}.

\bibitem[{K{\"o}pf et~al.(2023)K{\"o}pf, Kilcher, von R{\"u}tte, Anagnostidis, Tam, Stevens, Barhoum, Nguyen, Stanley, Nagyfi, ES, Suri, Glushkov, Dantuluri, Maguire, Schuhmann, Nguyen, and Mattick}]{kopf2023openassistant}
Andreas K{\"o}pf, Yannic Kilcher, Dimitri von R{\"u}tte, Sotiris Anagnostidis, Zhi~Rui Tam, Keith Stevens, Abdullah Barhoum, Duc~Minh Nguyen, Oliver Stanley, Rich{\'a}rd Nagyfi, Shahul ES, Sameer Suri, David~Alexandrovich Glushkov, Arnav~Varma Dantuluri, Andrew Maguire, Christoph Schuhmann, Huu Nguyen, and Alexander~Julian Mattick. 2023.
\newblock \href {https://openreview.net/forum?id=VSJotgbPHF} {Openassistant conversations - democratizing large language model alignment}.
\newblock In \emph{Thirty-seventh Conference on Neural Information Processing Systems Datasets and Benchmarks Track}.

\bibitem[{Li et~al.(2023)Li, Zhang, Dubois, Taori, Gulrajani, Guestrin, Liang, and Hashimoto}]{alpaca_eval}
Xuechen Li, Tianyi Zhang, Yann Dubois, Rohan Taori, Ishaan Gulrajani, Carlos Guestrin, Percy Liang, and Tatsunori~B. Hashimoto. 2023.
\newblock \href {https://github.com/tatsu-lab/alpaca_eval} {Alpacaeval: An automatic evaluator of instruction-following models}.
\newblock Github repository.

\bibitem[{Liang et~al.(2022)Liang, Bommasani, Lee, Tsipras, Soylu, Yasunaga, Zhang, Narayanan, Wu, Kumar et~al.}]{liang2022holistic}
Percy Liang, Rishi Bommasani, Tony Lee, Dimitris Tsipras, Dilara Soylu, Michihiro Yasunaga, Yian Zhang, Deepak Narayanan, Yuhuai Wu, Ananya Kumar, et~al. 2022.
\newblock \href {https://arxiv.org/abs/2211.09110} {Holistic evaluation of language models}.
\newblock \emph{arXiv preprint arXiv:2211.09110}.

\bibitem[{Lin et~al.(2022)Lin, Hilton, and Evans}]{lin2022truthfulqa}
Stephanie Lin, Jacob Hilton, and Owain Evans. 2022.
\newblock Truthfulqa: Measuring how models mimic human falsehoods.
\newblock In \emph{Proceedings of the 60th Annual Meeting of the Association for Computational Linguistics (Volume 1: Long Papers)}, pages 3214--3252.

\bibitem[{Liu et~al.(2020)Liu, Cui, Liu, Huang, Wang, and Zhang}]{logi-qa}
Jian Liu, Leyang Cui, Hanmeng Liu, Dandan Huang, Yile Wang, and Yue Zhang. 2020.
\newblock \href {http://arxiv.org/abs/2007.08124} {Logiqa: {A} challenge dataset for machine reading comprehension with logical reasoning}.
\newblock \emph{CoRR}, abs/2007.08124.

\bibitem[{Liu et~al.(2023)Liu, Qiao, Neiswanger, Wang, Tan, Tao, Li, Wang, Sun, Pangarkar et~al.}]{liu2023llm360}
Zhengzhong Liu, Aurick Qiao, Willie Neiswanger, Hongyi Wang, Bowen Tan, Tianhua Tao, Junbo Li, Yuqi Wang, Suqi Sun, Omkar Pangarkar, et~al. 2023.
\newblock \href {https://arxiv.org/abs/2312.06550} {Llm360: Towards fully transparent open-source llms}.
\newblock \emph{arXiv preprint arXiv:2312.06550}.

\bibitem[{Loshchilov and Hutter(2019)}]{loshchilov2018decoupled}
Ilya Loshchilov and Frank Hutter. 2019.
\newblock \href {https://openreview.net/forum?id=Bkg6RiCqY7} {Decoupled weight decay regularization}.
\newblock In \emph{International Conference on Learning Representations}.

\bibitem[{Luccioni et~al.(2022)Luccioni, Viguier, and Ligozat}]{luccioni2022estimating}
Alexandra~Sasha Luccioni, Sylvain Viguier, and Anne-Laure Ligozat. 2022.
\newblock \href {http://arxiv.org/abs/2211.02001} {Estimating the carbon footprint of bloom, a 176b parameter language model}.

\bibitem[{Magnusson et~al.(2023)Magnusson, Bhagia, Hofmann, Soldaini, Jha, Tafjord, Schwenk, Walsh, Elazar, Lo et~al.}]{magnusson2023paloma}
Ian Magnusson, Akshita Bhagia, Valentin Hofmann, Luca Soldaini, Ananya~Harsh Jha, Oyvind Tafjord, Dustin Schwenk, Evan~Pete Walsh, Yanai Elazar, Kyle Lo, et~al. 2023.
\newblock Paloma: A benchmark for evaluating language model fit.
\newblock \emph{arXiv preprint arXiv:2312.10523}.

\bibitem[{Marcus et~al.(1999)Marcus, Santorini, Marcinkiewicz, and Taylor}]{marcusptb}
Mitchell~P. Marcus, Beatrice Santorini, Mary~Ann Marcinkiewicz, and Ann Taylor. 1999.
\newblock \href {https://doi.org/10.35111/GQ1X-J780} {Treebank-3}.

\bibitem[{Merity et~al.(2016)Merity, Xiong, Bradbury, and Socher}]{merity2016pointersm}
Stephen Merity, Caiming Xiong, James Bradbury, and Richard Socher. 2016.
\newblock \href {https://api.semanticscholar.org/CorpusID:16299141} {Pointer sentinel mixture models}.
\newblock \emph{ArXiv}, abs/1609.07843.

\bibitem[{Micikevicius et~al.(2017)Micikevicius, Narang, Alben, Diamos, Elsen, Garc{\'i}a, Ginsburg, Houston, Kuchaiev, Venkatesh, and Wu}]{Micikevicius2017MixedPT}
Paulius Micikevicius, Sharan Narang, Jonah Alben, Gregory~Frederick Diamos, Erich Elsen, David Garc{\'i}a, Boris Ginsburg, Michael Houston, Oleksii Kuchaiev, Ganesh Venkatesh, and Hao Wu. 2017.
\newblock \href {https://api.semanticscholar.org/CorpusID:3297437} {Mixed precision training}.
\newblock \emph{ArXiv}, abs/1710.03740.

\bibitem[{Mihaylov et~al.(2018)Mihaylov, Clark, Khot, and Sabharwal}]{mihaylov2018can}
Todor Mihaylov, Peter Clark, Tushar Khot, and Ashish Sabharwal. 2018.
\newblock \href {https://arxiv.org/abs/1809.02789} {Can a suit of armor conduct electricity? a new dataset for open book question answering}.
\newblock \emph{arXiv preprint arXiv:1809.02789}.

\bibitem[{Mikolov et~al.(2013)Mikolov, Sutskever, Chen, Corrado, and Dean}]{Mikolov2013DistributedRO}
Tomas Mikolov, Ilya Sutskever, Kai Chen, Gregory~S. Corrado, and Jeffrey Dean. 2013.
\newblock \href {https://api.semanticscholar.org/CorpusID:16447573} {Distributed representations of words and phrases and their compositionality}.
\newblock In \emph{Neural Information Processing Systems}.

\bibitem[{Mishra et~al.(2022)Mishra, Khashabi, Baral, and Hajishirzi}]{mishra-etal-2022-cross}
Swaroop Mishra, Daniel Khashabi, Chitta Baral, and Hannaneh Hajishirzi. 2022.
\newblock \href {https://doi.org/10.18653/v1/2022.acl-long.244} {Cross-task generalization via natural language crowdsourcing instructions}.
\newblock In \emph{Proceedings of the 60th Annual Meeting of the Association for Computational Linguistics (Volume 1: Long Papers)}, pages 3470--3487, Dublin, Ireland. Association for Computational Linguistics.

\bibitem[{{MosaicML NLP Team}(2023)}]{MosaicML2023Introducing}
{MosaicML NLP Team}. 2023.
\newblock \href {https://www.mosaicml.com/blog/mpt-7b} {Introducing mpt-7b: A new standard for open-source, commercially usable llms}.
\newblock Accessed: 2023-05-05.

\bibitem[{Muennighoff et~al.(2023)Muennighoff, Rush, Barak, Scao, Piktus, Tazi, Pyysalo, Wolf, and Raffel}]{muennighoff2023scaling}
Niklas Muennighoff, Alexander~M Rush, Boaz Barak, Teven~Le Scao, Aleksandra Piktus, Nouamane Tazi, Sampo Pyysalo, Thomas Wolf, and Colin Raffel. 2023.
\newblock Scaling data-constrained language models.
\newblock \emph{arXiv preprint arXiv:2305.16264}.

\bibitem[{Nunes(2020)}]{nunesptb}
Davide Nunes. 2020.
\newblock \href {https://doi.org/10.5281/ZENODO.3910021} {Preprocessed penn tree bank}.

\bibitem[{OpenAI(2023)}]{OpenAI2023GPT4TR}
OpenAI. 2023.
\newblock \href {https://api.semanticscholar.org/CorpusID:257532815} {Gpt-4 technical report}.
\newblock \emph{ArXiv}, abs/2303.08774.

\bibitem[{Ouyang et~al.(2022)Ouyang, Wu, Jiang, Almeida, Wainwright, Mishkin, Zhang, Agarwal, Slama, Ray, Schulman, Hilton, Kelton, Miller, Simens, Askell, Welinder, Christiano, Leike, and Lowe}]{NEURIPS2022_b1efde53}
Long Ouyang, Jeffrey Wu, Xu~Jiang, Diogo Almeida, Carroll Wainwright, Pamela Mishkin, Chong Zhang, Sandhini Agarwal, Katarina Slama, Alex Ray, John Schulman, Jacob Hilton, Fraser Kelton, Luke Miller, Maddie Simens, Amanda Askell, Peter Welinder, Paul~F Christiano, Jan Leike, and Ryan Lowe. 2022.
\newblock \href {https://proceedings.neurips.cc/paper_files/paper/2022/file/b1efde53be364a73914f58805a001731-Paper-Conference.pdf} {Training language models to follow instructions with human feedback}.
\newblock In \emph{Advances in Neural Information Processing Systems}, volume~35, pages 27730--27744. Curran Associates, Inc.

\bibitem[{Papasavva et~al.(2020)Papasavva, Zannettou, Cristofaro, Stringhini, and Blackburn}]{papasavva_2020}
Antonis Papasavva, Savvas Zannettou, Emiliano~De Cristofaro, Gianluca Stringhini, and Jeremy Blackburn. 2020.
\newblock \href {https://doi.org/10.1609/icwsm.v14i1.7354} {Raiders of the lost kek: 3.5 years of augmented 4chan posts from the politically incorrect board}.
\newblock \emph{Proceedings of the International {AAAI} Conference on Web and Social Media}, 14:885--894.

\bibitem[{Patterson et~al.(2021)Patterson, Gonzalez, Le, Liang, Munguia, Rothchild, So, Texier, and Dean}]{patterson2021carbon}
David Patterson, Joseph Gonzalez, Quoc Le, Chen Liang, Lluis-Miquel Munguia, Daniel Rothchild, David So, Maud Texier, and Jeff Dean. 2021.
\newblock \href {http://arxiv.org/abs/2104.10350} {Carbon emissions and large neural network training}.

\bibitem[{Penedo et~al.(2023)Penedo, Malartic, Hesslow, Cojocaru, Cappelli, Alobeidli, Pannier, Almazrouei, and Launay}]{penedo2023therd}
Guilherme Penedo, Quentin Malartic, Daniel Hesslow, Ruxandra-Aim{\'e}e Cojocaru, Alessandro Cappelli, Hamza Alobeidli, Baptiste Pannier, Ebtesam Almazrouei, and Julien Launay. 2023.
\newblock \href {https://api.semanticscholar.org/CorpusID:259063761} {The refinedweb dataset for falcon llm: Outperforming curated corpora with web data, and web data only}.
\newblock \emph{ArXiv}, abs/2306.01116.

\bibitem[{Peters et~al.(2018)Peters, Neumann, Iyyer, Gardner, Clark, Lee, and Zettlemoyer}]{Peters2018DeepCW}
Matthew~E. Peters, Mark Neumann, Mohit Iyyer, Matt Gardner, Christopher Clark, Kenton Lee, and Luke Zettlemoyer. 2018.
\newblock \href {https://api.semanticscholar.org/CorpusID:3626819} {Deep contextualized word representations}.
\newblock \emph{ArXiv}, abs/1802.05365.

\bibitem[{Pilehvar and Camacho{-}Collados(2018)}]{wic}
Mohammad~Taher Pilehvar and Jos{\'{e}} Camacho{-}Collados. 2018.
\newblock \href {http://arxiv.org/abs/1808.09121} {Wic: 10, 000 example pairs for evaluating context-sensitive representations}.
\newblock \emph{CoRR}, abs/1808.09121.

\bibitem[{Rae et~al.(2022)Rae, Borgeaud, Cai, Millican, Hoffmann, Song, Aslanides, Henderson, Ring, Young, Rutherford, Hennigan, Menick, Cassirer, Powell, van~den Driessche, Hendricks, Rauh, Huang, Glaese, Welbl, Dathathri, Huang, Uesato, Mellor, Higgins, Creswell, McAleese, Wu, Elsen, Jayakumar, Buchatskaya, Budden, Sutherland, Simonyan, Paganini, Sifre, Martens, Li, Kuncoro, Nematzadeh, Gribovskaya, Donato, Lazaridou, Mensch, Lespiau, Tsimpoukelli, Grigorev, Fritz, Sottiaux, Pajarskas, Pohlen, Gong, Toyama, de~Masson~d'Autume, Li, Terzi, Mikulik, Babuschkin, Clark, de~Las~Casas, Guy, Jones, Bradbury, Johnson, Hechtman, Weidinger, Gabriel, Isaac, Lockhart, Osindero, Rimell, Dyer, Vinyals, Ayoub, Stanway, Bennett, Hassabis, Kavukcuoglu, and Irving}]{rae2022scaling}
Jack~W. Rae, Sebastian Borgeaud, Trevor Cai, Katie Millican, Jordan Hoffmann, Francis Song, John Aslanides, Sarah Henderson, Roman Ring, Susannah Young, Eliza Rutherford, Tom Hennigan, Jacob Menick, Albin Cassirer, Richard Powell, George van~den Driessche, Lisa~Anne Hendricks, Maribeth Rauh, Po-Sen Huang, Amelia Glaese, Johannes Welbl, Sumanth Dathathri, Saffron Huang, Jonathan Uesato, John Mellor, Irina Higgins, Antonia Creswell, Nat McAleese, Amy Wu, Erich Elsen, Siddhant Jayakumar, Elena Buchatskaya, David Budden, Esme Sutherland, Karen Simonyan, Michela Paganini, Laurent Sifre, Lena Martens, Xiang~Lorraine Li, Adhiguna Kuncoro, Aida Nematzadeh, Elena Gribovskaya, Domenic Donato, Angeliki Lazaridou, Arthur Mensch, Jean-Baptiste Lespiau, Maria Tsimpoukelli, Nikolai Grigorev, Doug Fritz, Thibault Sottiaux, Mantas Pajarskas, Toby Pohlen, Zhitao Gong, Daniel Toyama, Cyprien de~Masson~d'Autume, Yujia Li, Tayfun Terzi, Vladimir Mikulik, Igor Babuschkin, Aidan Clark, Diego de~Las~Casas, Aurelia Guy, Chris Jones,
  James Bradbury, Matthew Johnson, Blake Hechtman, Laura Weidinger, Iason Gabriel, William Isaac, Ed~Lockhart, Simon Osindero, Laura Rimell, Chris Dyer, Oriol Vinyals, Kareem Ayoub, Jeff Stanway, Lorrayne Bennett, Demis Hassabis, Koray Kavukcuoglu, and Geoffrey Irving. 2022.
\newblock \href {http://arxiv.org/abs/2112.11446} {Scaling language models: Methods, analysis \& insights from training gopher}.

\bibitem[{Rafailov et~al.(2023)Rafailov, Sharma, Mitchell, Manning, Ermon, and Finn}]{rafailov2023direct}
Rafael Rafailov, Archit Sharma, Eric Mitchell, Christopher~D Manning, Stefano Ermon, and Chelsea Finn. 2023.
\newblock \href {https://openreview.net/forum?id=HPuSIXJaa9} {Direct preference optimization: Your language model is secretly a reward model}.
\newblock In \emph{Thirty-seventh Conference on Neural Information Processing Systems}.

\bibitem[{Raffel et~al.(2020)Raffel, Shazeer, Roberts, Lee, Narang, Matena, Zhou, Li, and Liu}]{raffel2020exploring}
Colin Raffel, Noam Shazeer, Adam Roberts, Katherine Lee, Sharan Narang, Michael Matena, Yanqi Zhou, Wei Li, and Peter~J. Liu. 2020.
\newblock Exploring the limits of transfer learning with a unified text-to-text transformer.
\newblock \emph{J. Mach. Learn. Res.}, 21(1).

\bibitem[{Rajbhandari et~al.(2019)Rajbhandari, Rasley, Ruwase, and He}]{Rajbhandari2019ZeRO}
Samyam Rajbhandari, Jeff Rasley, Olatunji Ruwase, and Yuxiong He. 2019.
\newblock \href {https://api.semanticscholar.org/CorpusID:203736482} {Zero: Memory optimizations toward training trillion parameter models}.
\newblock \emph{SC20: International Conference for High Performance Computing, Networking, Storage and Analysis}, pages 1--16.

\bibitem[{Reid et~al.(2022)Reid, Zhong, Gururangan, and Zettlemoyer}]{reid-etal-2022-m2d2}
Machel Reid, Victor Zhong, Suchin Gururangan, and Luke Zettlemoyer. 2022.
\newblock \href {https://aclanthology.org/2022.emnlp-main.63} {{M}2{D}2: A massively multi-domain language modeling dataset}.
\newblock In \emph{Proceedings of the 2022 Conference on Empirical Methods in Natural Language Processing}, pages 964--975, Abu Dhabi, United Arab Emirates. Association for Computational Linguistics.

\bibitem[{Ribeiro et~al.(2021)Ribeiro, Blackburn, Bradlyn, Cristofaro, Stringhini, Long, Greenberg, and Zannettou}]{Horta_Ribeiro_2021}
Manoel~Horta Ribeiro, Jeremy Blackburn, Barry Bradlyn, Emiliano~De Cristofaro, Gianluca Stringhini, Summer Long, Stephanie Greenberg, and Savvas Zannettou. 2021.
\newblock \href {https://doi.org/10.1609/icwsm.v15i1.18053} {The evolution of the manosphere across the web}.
\newblock \emph{Proceedings of the International {AAAI} Conference on Web and Social Media}, 15:196--207.

\bibitem[{Rosenfeld(2000)}]{rosenfeld2000two}
Ronald Rosenfeld. 2000.
\newblock Two decades of statistical language modeling: Where do we go from here?
\newblock \emph{Proceedings of the IEEE}, 88(8):1270--1278.

\bibitem[{Sakaguchi et~al.(2021)Sakaguchi, Bras, Bhagavatula, and Choi}]{sakaguchi2021winogrande}
Keisuke Sakaguchi, Ronan~Le Bras, Chandra Bhagavatula, and Yejin Choi. 2021.
\newblock \href {https://dl.acm.org/doi/abs/10.1145/3474381} {Winogrande: An adversarial winograd schema challenge at scale}.
\newblock \emph{Communications of the ACM}, 64(9):99--106.

\bibitem[{Sanh et~al.(2022)Sanh, Webson, Raffel, Bach, Sutawika, Alyafeai, Chaffin, Stiegler, Raja, Dey, Bari, Xu, Thakker, Sharma, Szczechla, Kim, Chhablani, Nayak, Datta, Chang, Jiang, Wang, Manica, Shen, Yong, Pandey, Bawden, Wang, Neeraj, Rozen, Sharma, Santilli, Fevry, Fries, Teehan, Scao, Biderman, Gao, Wolf, and Rush}]{sanh2022multitask}
Victor Sanh, Albert Webson, Colin Raffel, Stephen Bach, Lintang Sutawika, Zaid Alyafeai, Antoine Chaffin, Arnaud Stiegler, Arun Raja, Manan Dey, M~Saiful Bari, Canwen Xu, Urmish Thakker, Shanya~Sharma Sharma, Eliza Szczechla, Taewoon Kim, Gunjan Chhablani, Nihal Nayak, Debajyoti Datta, Jonathan Chang, Mike Tian-Jian Jiang, Han Wang, Matteo Manica, Sheng Shen, Zheng~Xin Yong, Harshit Pandey, Rachel Bawden, Thomas Wang, Trishala Neeraj, Jos Rozen, Abheesht Sharma, Andrea Santilli, Thibault Fevry, Jason~Alan Fries, Ryan Teehan, Teven~Le Scao, Stella Biderman, Leo Gao, Thomas Wolf, and Alexander~M Rush. 2022.
\newblock \href {https://openreview.net/forum?id=9Vrb9D0WI4} {Multitask prompted training enables zero-shot task generalization}.
\newblock In \emph{International Conference on Learning Representations}.

\bibitem[{Shazeer(2020)}]{Shazeer2020GLUVI}
Noam~M. Shazeer. 2020.
\newblock \href {https://api.semanticscholar.org/CorpusID:211096588} {Glu variants improve transformer}.
\newblock \emph{ArXiv}, abs/2002.05202.

\bibitem[{Soldaini et~al.(2024)Soldaini, Kinney, Bhagia, Schwenk, Atkinson, Authur, Bogin, Chandu, Dumas, Elazar, Hofmann, Jha, Kumar, Lucy, Lyu, Lambert, Magnusson, Morrison, Muennighoff, Naik, Nam, Peters, Ravichander, Richardson, Shen, Strubell, Subramani, Tafjord, Walsh, Zettlemoyer, Smith, Hajishirzi, Beltagy, Groeneveld, Dodge, and Lo}]{dolma}
Luca Soldaini, Rodney Kinney, Akshita Bhagia, Dustin Schwenk, David Atkinson, Russell Authur, Ben Bogin, Khyathi Chandu, Jennifer Dumas, Yanai Elazar, Valentin Hofmann, Ananya~Harsh Jha, Sachin Kumar, Li~Lucy, Xinxi Lyu, Nathan Lambert, Ian Magnusson, Jacob Morrison, Niklas Muennighoff, Aakanksha Naik, Crystal Nam, Matthew~E. Peters, Abhilasha Ravichander, Kyle Richardson, Zejiang Shen, Emma Strubell, Nishant Subramani, Oyvind Tafjord, Pete Walsh, Luke Zettlemoyer, Noah~A. Smith, Hannaneh Hajishirzi, Iz~Beltagy, Dirk Groeneveld, Jesse Dodge, and Kyle Lo. 2024.
\newblock {Dolma: an Open Corpus of Three Trillion Tokens for Language Model Pretraining Research}.
\newblock \emph{arXiv preprint}.

\bibitem[{Strubell et~al.(2019)Strubell, Ganesh, and McCallum}]{strubell-etal-2019-energy}
Emma Strubell, Ananya Ganesh, and Andrew McCallum. 2019.
\newblock \href {https://doi.org/10.18653/v1/P19-1355} {Energy and policy considerations for deep learning in {NLP}}.
\newblock In \emph{Proceedings of the 57th Annual Meeting of the Association for Computational Linguistics}, pages 3645--3650, Florence, Italy. Association for Computational Linguistics.

\bibitem[{Su et~al.(2021)Su, Lu, Pan, Wen, and Liu}]{Su2021RoFormerET}
Jianlin Su, Yu~Lu, Shengfeng Pan, Bo~Wen, and Yunfeng Liu. 2021.
\newblock \href {https://api.semanticscholar.org/CorpusID:233307138} {Roformer: Enhanced transformer with rotary position embedding}.
\newblock \emph{ArXiv}, abs/2104.09864.

\bibitem[{Taori et~al.(2023)Taori, Gulrajani, Zhang, Dubois, Li, Guestrin, Liang, and Hashimoto}]{alpaca}
Rohan Taori, Ishaan Gulrajani, Tianyi Zhang, Yann Dubois, Xuechen Li, Carlos Guestrin, Percy Liang, and Tatsunori~B. Hashimoto. 2023.
\newblock Stanford alpaca: An instruction-following llama model.
\newblock \url{https://github.com/tatsu-lab/stanford_alpaca}.

\bibitem[{Teknium1(2023)}]{gpteacher}
Teknium1. 2023.
\newblock Gpteacher.
\newblock \url{https://github.com/teknium1/GPTeacher}.

\bibitem[{{Together Computer}(2023)}]{together2023redpajama}
{Together Computer}. 2023.
\newblock \href {https://github.com/togethercomputer/RedPajama-Data} {{RedPajama: An Open Source Recipe to Reproduce LLaMA training dataset}}.

\bibitem[{Touvron et~al.(2023{\natexlab{a}})Touvron, Lavril, Izacard, Martinet, Lachaux, Lacroix, Rozi{\`e}re, Goyal, Hambro, Azhar, Rodriguez, Joulin, Grave, and Lample}]{touvron2023llama1}
Hugo Touvron, Thibaut Lavril, Gautier Izacard, Xavier Martinet, Marie-Anne Lachaux, Timoth{\'e}e Lacroix, Baptiste Rozi{\`e}re, Naman Goyal, Eric Hambro, Faisal Azhar, Aurelien Rodriguez, Armand Joulin, Edouard Grave, and Guillaume Lample. 2023{\natexlab{a}}.
\newblock \href {https://api.semanticscholar.org/CorpusID:257219404} {Llama: Open and efficient foundation language models}.
\newblock \emph{ArXiv}, abs/2302.13971.

\bibitem[{Touvron et~al.(2023{\natexlab{b}})Touvron, Martin, Stone, Albert, Almahairi, Babaei, Bashlykov, Batra, Bhargava, Bhosale, Bikel, Blecher, Ferrer, Chen, Cucurull, Esiobu, Fernandes, Fu, Fu, Fuller, Gao, Goswami, Goyal, Hartshorn, Hosseini, Hou, Inan, Kardas, Kerkez, Khabsa, Kloumann, Korenev, Koura, Lachaux, Lavril, Lee, Liskovich, Lu, Mao, Martinet, Mihaylov, Mishra, Molybog, Nie, Poulton, Reizenstein, Rungta, Saladi, Schelten, Silva, Smith, Subramanian, Tan, Tang, Taylor, Williams, Kuan, Xu, Yan, Zarov, Zhang, Fan, Kambadur, Narang, Rodriguez, Stojnic, Edunov, and Scialom}]{touvron2023llama2}
Hugo Touvron, Louis Martin, Kevin Stone, Peter Albert, Amjad Almahairi, Yasmine Babaei, Nikolay Bashlykov, Soumya Batra, Prajjwal Bhargava, Shruti Bhosale, Dan Bikel, Lukas Blecher, Cristian~Canton Ferrer, Moya Chen, Guillem Cucurull, David Esiobu, Jude Fernandes, Jeremy Fu, Wenyin Fu, Brian Fuller, Cynthia Gao, Vedanuj Goswami, Naman Goyal, Anthony Hartshorn, Saghar Hosseini, Rui Hou, Hakan Inan, Marcin Kardas, Viktor Kerkez, Madian Khabsa, Isabel Kloumann, Artem Korenev, Punit~Singh Koura, Marie-Anne Lachaux, Thibaut Lavril, Jenya Lee, Diana Liskovich, Yinghai Lu, Yuning Mao, Xavier Martinet, Todor Mihaylov, Pushkar Mishra, Igor Molybog, Yixin Nie, Andrew Poulton, Jeremy Reizenstein, Rashi Rungta, Kalyan Saladi, Alan Schelten, Ruan Silva, Eric~Michael Smith, Ranjan Subramanian, Xiaoqing~Ellen Tan, Binh Tang, Ross Taylor, Adina Williams, Jian~Xiang Kuan, Puxin Xu, Zheng Yan, Iliyan Zarov, Yuchen Zhang, Angela Fan, Melanie Kambadur, Sharan Narang, Aurelien Rodriguez, Robert Stojnic, Sergey Edunov, and Thomas
  Scialom. 2023{\natexlab{b}}.
\newblock \href {http://arxiv.org/abs/2307.09288} {Llama 2: Open foundation and fine-tuned chat models}.

\bibitem[{Ubierna et~al.(2022)Ubierna, Santos, and Mercier-Blais}]{Ubierna2022}
Mar{\'i}a Ubierna, Cristina~D{\'i}ez Santos, and Sara Mercier-Blais. 2022.
\newblock \href {https://doi.org/10.1007/978-981-16-5493-0_5} {\emph{Water Security and Climate Change: Hydropower Reservoir Greenhouse Gas Emissions}}, pages 69--94. Springer Singapore, Singapore.

\bibitem[{Vaswani et~al.(2017)Vaswani, Shazeer, Parmar, Uszkoreit, Jones, Gomez, Kaiser, and Polosukhin}]{vaswani2017attention}
Ashish Vaswani, Noam Shazeer, Niki Parmar, Jakob Uszkoreit, Llion Jones, Aidan~N Gomez, \L~ukasz Kaiser, and Illia Polosukhin. 2017.
\newblock \href {https://proceedings.neurips.cc/paper_files/paper/2017/file/3f5ee243547dee91fbd053c1c4a845aa-Paper.pdf} {Attention is all you need}.
\newblock In \emph{Advances in Neural Information Processing Systems}, volume~30. Curran Associates, Inc.

\bibitem[{Vilares and G{\'o}mez-Rodr{\'\i}guez(2019)}]{head-qa}
David Vilares and Carlos G{\'o}mez-Rodr{\'\i}guez. 2019.
\newblock \href {https://doi.org/10.18653/v1/P19-1092} {{HEAD}-{QA}: A healthcare dataset for complex reasoning}.
\newblock In \emph{Proceedings of the 57th Annual Meeting of the Association for Computational Linguistics}, pages 960--966, Florence, Italy. Association for Computational Linguistics.

\bibitem[{Wang et~al.(2018)Wang, Singh, Michael, Hill, Levy, and Bowman}]{glue}
Alex Wang, Amanpreet Singh, Julian Michael, Felix Hill, Omer Levy, and Samuel~R. Bowman. 2018.
\newblock \href {https://arxiv.org/abs/1804.07461} {Glue: A multi-task benchmark and analysis platform for natural language understanding}.
\newblock \emph{ArXiv}, abs/1804.07461.

\bibitem[{Wang et~al.(2023)Wang, Ivison, Dasigi, Hessel, Khot, Chandu, Wadden, MacMillan, Smith, Beltagy, and Hajishirzi}]{wang2023far}
Yizhong Wang, Hamish Ivison, Pradeep Dasigi, Jack Hessel, Tushar Khot, Khyathi~Raghavi Chandu, David Wadden, Kelsey MacMillan, Noah~A. Smith, Iz~Beltagy, and Hannaneh Hajishirzi. 2023.
\newblock \href {http://arxiv.org/abs/2306.04751} {How far can camels go? exploring the state of instruction tuning on open resources}.

\bibitem[{Wei et~al.(2022)Wei, Bosma, Zhao, Guu, Yu, Lester, Du, Dai, and Le}]{wei2022finetuned}
Jason Wei, Maarten Bosma, Vincent Zhao, Kelvin Guu, Adams~Wei Yu, Brian Lester, Nan Du, Andrew~M. Dai, and Quoc~V Le. 2022.
\newblock \href {https://openreview.net/forum?id=gEZrGCozdqR} {Finetuned language models are zero-shot learners}.
\newblock In \emph{International Conference on Learning Representations}.

\bibitem[{Welbl et~al.(2017)Welbl, Liu, and Gardner}]{welbl2017crowdsourcing}
Johannes Welbl, Nelson~F Liu, and Matt Gardner. 2017.
\newblock \href {https://arxiv.org/abs/1707.06209} {Crowdsourcing multiple choice science questions}.
\newblock \emph{arXiv preprint arXiv:1707.06209}.

\bibitem[{Wu et~al.(2022)Wu, Raghavendra, Gupta, Acun, Ardalani, Maeng, Chang, Behram, Huang, Bai, Gschwind, Gupta, Ott, Melnikov, Candido, Brooks, Chauhan, Lee, Lee, Akyildiz, Balandat, Spisak, Jain, Rabbat, and Hazelwood}]{wu2022sustainable}
Carole-Jean Wu, Ramya Raghavendra, Udit Gupta, Bilge Acun, Newsha Ardalani, Kiwan Maeng, Gloria Chang, Fiona~Aga Behram, James Huang, Charles Bai, Michael Gschwind, Anurag Gupta, Myle Ott, Anastasia Melnikov, Salvatore Candido, David Brooks, Geeta Chauhan, Benjamin Lee, Hsien-Hsin~S. Lee, Bugra Akyildiz, Maximilian Balandat, Joe Spisak, Ravi Jain, Mike Rabbat, and Kim Hazelwood. 2022.
\newblock \href {http://arxiv.org/abs/2111.00364} {Sustainable ai: Environmental implications, challenges and opportunities}.

\bibitem[{Xu et~al.(2024)Xu, Sun, Zheng, Geng, Zhao, Feng, Tao, Lin, and Jiang}]{xu2024wizardlm}
Can Xu, Qingfeng Sun, Kai Zheng, Xiubo Geng, Pu~Zhao, Jiazhan Feng, Chongyang Tao, Qingwei Lin, and Daxin Jiang. 2024.
\newblock \href {https://openreview.net/forum?id=CfXh93NDgH} {Wizard{LM}: Empowering large pre-trained language models to follow complex instructions}.
\newblock In \emph{The Twelfth International Conference on Learning Representations}.

\bibitem[{Xu et~al.(2023)Xu, Guo, Duan, and McAuley}]{xu2023baize}
Canwen Xu, Daya Guo, Nan Duan, and Julian McAuley. 2023.
\newblock Baize: An open-source chat model with parameter-efficient tuning on self-chat data.
\newblock \emph{arXiv preprint arXiv:2304.01196}.

\bibitem[{Zannettou et~al.(2018)Zannettou, Bradlyn, De~Cristofaro, Kwak, Sirivianos, Stringini, and Blackburn}]{zannettougab}
Savvas Zannettou, Barry Bradlyn, Emiliano De~Cristofaro, Haewoon Kwak, Michael Sirivianos, Gianluca Stringini, and Jeremy Blackburn. 2018.
\newblock \href {https://doi.org/10.1145/3184558.3191531} {What is gab: A bastion of free speech or an alt-right echo chamber}.
\newblock In \emph{Companion Proceedings of the The Web Conference 2018}, WWW '18, page 1007–1014, Republic and Canton of Geneva, CHE. International World Wide Web Conferences Steering Committee.

\bibitem[{Zellers et~al.(2019)Zellers, Holtzman, Bisk, Farhadi, and Choi}]{zellers2019hellaswag}
Rowan Zellers, Ari Holtzman, Yonatan Bisk, Ali Farhadi, and Yejin Choi. 2019.
\newblock \href {https://arxiv.org/abs/1905.07830} {Hellaswag: Can a machine really finish your sentence?}
\newblock \emph{arXiv preprint arXiv:1905.07830}.

\bibitem[{Zhang and Sennrich(2019)}]{RMSNorm}
Biao Zhang and Rico Sennrich. 2019.
\newblock \href {https://api.semanticscholar.org/CorpusID:113405151} {Root mean square layer normalization}.
\newblock \emph{ArXiv}, abs/1910.07467.

\bibitem[{Zhang et~al.(2022)Zhang, Roller, Goyal, Artetxe, Chen, Chen, Dewan, Diab, Li, Lin, Mihaylov, Ott, Shleifer, Shuster, Simig, Koura, Sridhar, Wang, and Zettlemoyer}]{zhang2022opt}
Susan Zhang, Stephen Roller, Naman Goyal, Mikel Artetxe, Moya Chen, Shuohui Chen, Christopher Dewan, Mona Diab, Xian Li, Xi~Victoria Lin, Todor Mihaylov, Myle Ott, Sam Shleifer, Kurt Shuster, Daniel Simig, Punit~Singh Koura, Anjali Sridhar, Tianlu Wang, and Luke Zettlemoyer. 2022.
\newblock \href {http://arxiv.org/abs/2205.01068} {Opt: Open pre-trained transformer language models}.

\bibitem[{Zhao et~al.(2023)Zhao, Gu, Varma, Luo, chin Huang, Xu, Wright, Shojanazeri, Ott, Shleifer, Desmaison, Balioglu, Nguyen, Chauhan, Hao, and Li}]{Zhao2023PyTorchFSDP}
Yanli Zhao, Andrew Gu, Rohan Varma, Liangchen Luo, Chien chin Huang, Min Xu, Less Wright, Hamid Shojanazeri, Myle Ott, Sam Shleifer, Alban Desmaison, Can Balioglu, Bernard Nguyen, Geeta Chauhan, Yuchen Hao, and Shen Li. 2023.
\newblock \href {https://api.semanticscholar.org/CorpusID:258297871} {Pytorch fsdp: Experiences on scaling fully sharded data parallel}.
\newblock \emph{Proc. VLDB Endow.}, 16:3848--3860.

\end{thebibliography}
